\documentclass{article} 
\pdfoutput=1 
\usepackage{nips14submit_e,times}
\def \longversion

\usepackage{url}
\usepackage{amsmath,amsfonts,amssymb}
\usepackage{subfig}
\usepackage{mathtools}
\usepackage{amsopn}
\usepackage{dsfont}
\usepackage{wrapfig}
\usepackage{subfig}

\newcommand{\longtext}[1]{#1}
\newcommand{\shorttext}[1]{}
\usepackage{appendix}

\nipsfinalcopy

\usepackage{color}
\definecolor{forestgreenweb}{rgb}{0.13, 0.55, 0.13}
\newcommand{\cjl}[1]{}

\newcommand{\defeq}{\overset{\text{def}}{=}}
\newcommand{\logpropto}{\stackrel{{\scriptscriptstyle +}\mathsf{c}}{=}}

\newcommand{\setR}{\mathbb{R}}

\DeclareMathOperator*{\argmin}{arg\,min}

\def\bW{{\bf W}}

\def\bH{{\bf H}}

\def\bM{{\bf M}}
\def\bS{{\bf S}}
\def\bLambda{{\bf \Lambda}}

\usepackage{stmaryrd}
\newcommand{\atob}[2]{#1 \shortrightarrow #2}
\newcommand{\msga}[3]{
\mathchoice
  {{#1}_{\atob{#2}{#3}}}
 {{#1}_{\atob{#2}{#3}}}
 {{#1}_{\atob{#2}{#3}}}
 {{#1}_{\atob{#2}{#3}}}
} 

\newcommand{\msg}[2]{\msga{m}{#1}{#2}}
\newcommand{\msgax}[4]{
\mathchoice
 { {#1}_{\atob{#2}{#3}}^{#4}}
 { {#1}_{\atob{#2}{#3}}^{#4}}
 { {#1}_{\atob{#2}{#3}}^{#4}}
 { {#1}_{\atob{#2}{#3}}^{#4}}
} 
\newcommand{\msgx}[2]{\msgax{m}{#1}{#2}}

\title{Deep Unfolding:\\Model-Based Inspiration of Novel Deep Architectures}
\author{
John R. Hershey\\
MERL\\
Cambridge, MA, USA\\
\texttt{hershey@merl.com} \\
\And
Jonathan Le Roux\\
MERL\\
Cambridge, MA, USA\\
\texttt{leroux@merl.com} 
\And
Felix Weninger \\
TUM \\
Munich, Germany \\
\texttt{weninger@tum.de} \\
}

\begin{document}

\maketitle

\begin{abstract}
Model-based methods and deep neural networks have both been tremendously successful paradigms in machine learning. In model-based methods, problem domain knowledge can be built into the constraints of the model, typically at the expense of difficulties during inference. In contrast, deterministic deep neural networks are constructed in such a way that inference is straightforward, but their architectures are generic and it is unclear how to incorporate knowledge. This work aims to obtain the advantages of both approaches. To do so, we start with a model-based approach and an associated inference algorithm, and \emph{unfold} the inference iterations as layers in a deep network.  Rather than optimizing the original model, we \emph{untie} the model parameters across layers, in order to create a more powerful network.   The resulting architecture can be trained discriminatively to perform accurate inference within a fixed network size.  We show how this framework allows us to interpret conventional networks as mean-field inference in Markov random fields, and to obtain new architectures by instead using belief propagation as the inference algorithm.  We then show its application to a non-negative matrix factorization model that incorporates the problem-domain knowledge that sound sources are additive.  Deep unfolding of this model yields a new kind of non-negative deep neural network, that can be trained using a multiplicative backpropagation-style update algorithm. We present speech enhancement experiments showing that our approach is competitive with conventional neural networks despite using far fewer parameters.
\end{abstract}

\section{Introduction}
Two of the most successful general approaches in machine learning are model-based methods and deep neural networks (DNNs).  Each offers important well-known advantages and disadvantages.   The goal of this paper is to provide a general strategy to obtain the advantages of both approaches while avoiding many of their disadvantages.  The general idea can be summarized as follows:  given a model-based approach that requires an iterative inference method, we \emph{unfold} the iterations into a layer-wise structure analogous to a neural network.  We then \emph{untie} the model parameters across layers to obtain novel neural-network-like architectures that can easily be trained discriminatively using gradient-based methods.  The resulting formula combines the expressive power of a conventional deep network  with the internal structure of the model-based approach, while allowing inference to be performed in a fixed number of layers that can be optimized for best performance. We call this approach \emph{deep unfolding}.

A chief advantage of generative model-based approaches, such as probabilistic graphical models, is that they allow us to use prior knowledge and intuition to reason at the \emph{problem level} in devising inference algorithms, or what David Marr called the ``computational theory'' level of analysis \cite{marr1982vision,hershey2005perceptual}.  Important assumptions about problem constraints can often be incorporated into a model-based approach in a straightforward way.   Examples include constraints from the world, such as the linear additivity of signals, visual occlusion, three-dimensional geometry, as well as more subtle statistical assumptions such as conditional independence, latent variable structure, sparsity, low-rank covariances, and so on.   By hypothesizing and testing different problem-level constraints, insight into the nature of the problem can be gained and used as inspiration to improve the modeling assumptions \cite{hershey2005perceptual}.  Unfortunately, inference in probabilistic models can be both mathematically and computationally intractable.  Approximate methods, such as belief propagation and variational approximations, allow us to derive iterative algorithms to infer the latent variables of interest.  However, despite greatly improving the situation, such iterative methods are still often too slow for time-sensitive applications.  In such cases, rigorous discriminative optimization of such models can be challenging because they may involve bi-level optimization, where the parameter optimization depends in turn on an iterative inference algorithm \cite{colson2007overview}.

Deterministic deep neural networks, which have recently become the state of the art in many applications, are formulated such that the inference is defined as a finite closed-form expression, organized into \emph{layers} which are typically executed in sequence.  Discriminative training of the networks can be used to optimize the speed versus accuracy trade-off, and has become indispensable in producing systems that perform very well in a particular application. However, a well-known disadvantage is that conventional DNNs are closer to mechanisms than problem-level formulations, and can be considered essentially ``black-box'' methods.  It is difficult to incorporate prior knowledge about the world or the problem.  Moreover, even when one has a working DNN system, it is not clear how it actually achieves its results, and so discovering how to modify its architecture to achieve better results could be considered as much an art as a science.

The proposed methodology addresses these difficulties by bringing the problem-level formulation of model-based methods to the task of designing deep neural network architectures.  Each step of the process can be solved by well-known methods:  deriving iterative inference methods for a given probabilistic model follows a long tradition that makes use of many well-known tools,  and unfolding the iterations and applying the chain rule for gradient-based training is also relatively straightforward.

In the remainder of the paper we present a general framework that can be applied to a arbitrary models and inference algorithms.  
We then apply this framework to the case of inference in generic Markov random fields (MRFs), showing first how unfolding mean field inference in binary MRFs leads to conventional sigmoid networks, and how in contrast unfolding belief propagation leads to an alternative deep architecture.  We show how the two architectures can be unified and generalized using a power-mean formulation.   
We then focus on
a generative model of sound that embodies the problem-level assumption that signals mix linearly and therefore their power spectra are approximately additive.    

Despite the simplicity of its assumptions, this model, based on non-negative matrix factorization (NMF) \cite{Lee01-AFN,Smaragdis07-SAS,DNMF}, has no closed-form solution and relies on iterative inference methods, typically formulated as multiplicative updates.   A novel non-negative deep network architecture results from unfolding the iterations and untying the parameters.   This architecture can be more powerful than the original model-based NMF method, while still incorporating the basic additivity assumption from its problem-level analysis.  To optimize its non-negative parameters, we show how a new form of back-propagation, based on multiplicative updates, can be used to preserve non-negativity, without the need for constrained optimization.

Finally, we present experiments in the domain of speech enhancement, using the unfolded NMF model, showing that it is competitive in terms of accuracy with conventional sigmoid DNNs, while requiring a tenth of the number of parameters.

\textbf{Relationship to the literature}:
Some recent work has addressed the idea of unfolding inference algorithms and using gradient descent to optimize them in the context of a variety of models and inference methods.   Both sparse coding \cite{gregor2010learning, sprechmann2013supervised}, and non-negative matrix factorization \cite{yakarbilevel, sprechmann2014hscma} have been addressed using unfolding and back-propagation or other discriminative optimization methods.   In \cite{stoyanov2011empirical}, gradient-based optimization of loopy belief propagation was applied to binary pair-wise Markov random fields.  
In \cite{domke2011parameter, domke2013learning}, tree-reweighted belief propagation (TRW-BP) and mean field inference were unrolled and trained via gradient descent.  
In \cite{goodfellow2013multi} inference in a graphical model is implemented via an ensemble of unfolded inference algorithms trained to predict one held out variable given the others.   In all of these works, unfolding was done without untying parameters, so only an approximation to the original model was optimized.  

In our view, the untying of the parameters is the critical step in creating new deep architectures that can be competitive with conventional deep networks.  
\longtext{In a recent scene labeling paper \cite{ross2011learning}, learning of belief propagation and mean field message passing algorithms was considered for conditional random fields (CRFs), which are MRFs where all units are either inputs or labels.  In the case of belief propagation, rather than following the belief-propagation updates, all updates were replaced with standard logistic regression, which yields only conventional sigmoid networks, rather than novel architectures.  Simultaneously to our work, \cite{icml14_mfn} also introduced a similar untying approach applied to mean-field inference methods in MRFs.  Again, this results in conventional sigmoid networks, whose graph structure is derived from a schedule of mean field updates.  As of this writing there have been no novel deep architectures developed by unfolding and untying.  }
\shorttext{Some recent work has begun to address the untying of parameters for Markov random field inference algorithms.  Belief propagation-style inference was learned in \cite{ross2011learning}, using logistic regression with untied parameters.  Simultaneous to our work,  \cite{icml14_mfn} unfolded mean-field inference and untied parameters.  However, in both of these, only conventional sigmoid networks resulted from the untying.   \footnote{In Appendix \ref{sec:application-mrf}, we show how Markov random fields can be unfolded to form either classical sigmoid networks under mean-field inference or novel networks under belief propagation.  However, MRFs are generic models that do not rely on problem domain knowledge, an instead we focus in this paper on models that do.}}

\textbf{Main contributions of this paper}:  The novel contributions of this paper include: a framework for deriving novel deep network architectures from  model-based inference algorithms by unfolding the steps of the algorithm and untying the model parameters across iterations;  
\longtext{a generalization of classical sigmoid networks based on unification of different unfolded MRF inference algorithms; }
a novel non-negative deep network with non-negative parameters; a multiplicative form of back-propagation updates for training non-negative deep networks; finally, experiments showing the benefit of this approach in the domain of speech enhancement.

\section{General formulation of deep unfolding}
In the general setting, we consider models for which inference is an optimization problem.  One example is variational inference, where a lower bound on the data likelihood is optimized to estimate approximate posterior probabilities, which can then be used to compute conditional expected values of hidden variables. As another example, loopy belief propagation is an iterative algorithm that enforces local consistency constraints on marginal posteriors.  When it converges, the fixed points correspond to stationary points of the Bethe free energy \cite{yedidia2005constructing}. Finally, non-negative matrix factorization is a non-negative basis  expansion model, whose objective function can be optimized by simple multiplicative update rules.

Here, we present a general formulation based on a model, determined by \emph{parameters} $\theta$, that specifies the relationships between \emph{hidden} quantities of interest $y_i$ and the \emph{observed} variables $x_i$ for each data instance $i$.   The parameter set, $\theta$, contains all parameters used in the model:  for Markov random fields, $\theta$ contains the potential functions, for Gaussian-based models, it contains the means and variances, and for basis expansion models, it contains the basis functions.   The quantities of interest, $y_i$,  are typically estimates of latent variables important for a particular task.  For example, in a scene labeling task, $y_i$ might be the labels of the pixels; in denoising, $y_i$ might be the posterior mean of the latent clean signal.  At test time, estimating these quantities of interest involves optimizing an inference objective function $\mathcal{F}_{\theta}(x_i,\phi_i)$, where $\phi_i$ are intermediate variables (considered as vectors) from which $y_i$ can be computed:\footnote{We arbitrarily formulate it as a minimization, as in the case of energy minimization, but equivalently it could be a maximization as in the case of probabilities.}
\begin{align}
\hat{\phi}_i(x_i | \theta) = \arg \min_{\phi_i} \mathcal{F}_{\theta}(x_i,\phi_i),
\quad \hat{y}_i(x_i | \theta) = g_{\theta}(x_i,\hat{\phi}_i(x_i | \theta)),
\label{eq:general:obj1}
\end{align}
where $g_{\theta}$ is an estimator for $y_i$.
For many interesting cases, this optimization cannot be easily done and leads to an iterative inference algorithm. In probabilistic generative models, $\mathcal{F}$ might be an approximation to the negative log likelihood,  $y_i$ could be taken to represent hidden variables and $\phi_i$ to represent an estimate of their posterior distribution.
For example, in variational inference algorithms, $\phi_i$ could be taken to be the variational parameters. In sum-product loopy belief propagation, the $\phi_i$ would be the posterior marginal probabilities. On the other hand, for the non-probabilistic formulation of NMF, $\phi_i$ can be taken as the activation coefficients of the basis functions that are updated at inference time.
Note that the $x_i, y_i$ can all be sequences or have other underlying structure, but here for simplicity we ignore their structure.

At training time, we may optimize the parameters $\theta$ using a discriminative objective function,
\begin{align}
\mathcal{E}_{\theta} \defeq \sum_{i} \mathcal{D}(y^{*}_{i}, \hat{y}_i(x_i|\theta)),
\label{eq:general:obj2}
\end{align}
where $\mathcal{D}$ is a loss function and $y^{*}_{i}$ a reference value. In some settings, we can also consider a discriminative objective  $ \mathcal{D}(y^{*}_{i}, \hat{\phi}_i(x_i|\theta))$ which computes an expected loss.
 In the general case, minimization of \eqref{eq:general:obj2} is a \emph{bi-level} optimization problem since $\hat{y}_i(x_i | \theta)$ is itself determined by an optimization problem  \eqref{eq:general:obj1} that depends on the parameters $\theta$.

We assume that the intermediate variables $\phi_i$ in \eqref{eq:general:obj1} can be optimized iteratively using update steps $k \in \{1 \ldots K\}$ of the form,\footnote{Indices $k$ in superscript always refer to iteration index (similarly for $l$ defined later as the source index).}
\begin{align}
\phi^{k}_{i} = f_{\theta}(x_i,\phi^{k-1}_i),
\label{eq:general:update1}
\end{align}
beginning with $\phi^{0}_i$.
 Note that, although all steps are assumed to use the same $f_\theta$, it may be composed of smaller steps, each of which are different. This can occur in loopy belief propagation, when different messages are passed in each step, or in variational inference, when different variational parameters are updated in each step.

Although various optimization methods may be used, one of the simplest and most flexible involves applying gradient descent through the sequence of iterations of the approximate inference algorithm to optimize the parameters.  To make this tractable, and to match the training-time inference algorithm to the test-time procedure, the iterations can be truncated in the same way.   

However, rather than considering the iterations as an algorithm, we consider unfolding it as a sequence of layers in a neural network-like architecture, where the iteration index is now interpreted as an index to the neural network layer. The intermediate variables $\phi^{1}, \ldots, \phi^{K}$ are the nodes of layers $1$ to $K$ and \eqref{eq:general:update1} determines the transformation and activation function between layers. Finally, the ${y}^{K}_i$ are the nodes of the output layer, and are obtained by ${y}^{K}_i = g_{\theta}(x_i,\phi^{K}_i)$.
In the unfolded model presented so far, the parameters $\theta$ are tied across layers, and optimizing them is no different from optimizing the original model.

Although discriminative optimization can help in overcoming the mismatch between the assumed model and the data, it may not lead to fundamentally different model performance, and in particular may not be competitive with deep networks unless the model we start with is already deep.  However, in the deep unfolding framework, we hypothesize that a more powerful model can be obtained  by explicitly \emph{untying} the parameters across layers to allow the network to embody a more complex range of inference functions than the original model, with the original class of inference representing a special case.  Of course the cost of untying is the possibility of over-fitting, but this is the same situation for all deep neural networks, and can be handled in similar ways.  

To formulate this untying, we define parameters $\theta \defeq \{\theta^{k}\}_{k=0}^{K}$ for each layer, so that $\phi^{k}_{i} = f_{\theta^{k-1}}(x_i,\phi^{k-1}_i)$ and ${y}^{K}_i =  g_{\theta^{K}}(x_i,\phi^{K}_i)$.
Then we can compute the derivatives recursively as in back-propagation,
\begin{align}
\frac{\partial  \mathcal{E}}{\partial \phi^{K}_i}
&=  \frac{\partial  \mathcal{D}}{\partial {y}_i^{K}}\frac{\partial  {y}_i^{K}}{\partial \phi^{K}_i}
, &
\frac{\partial  \mathcal{E}}{\partial \theta^{K}_{\phantom{i}}}  &= \sum_i \frac{\partial  \mathcal{D}}{\partial {y}_i^{K}}\frac{\partial  {y}_i^{K}}{\partial \theta^{K}_{\phantom{i}}}
, \\
\frac{\partial  \mathcal{E}}{\partial \phi^{k}_i}  &= \frac{\partial  \mathcal{E}}{\partial \phi^{k+1}_i} \frac{\partial  \phi^{k+1}_i}{\partial \phi^{k}_i}
, &
\frac{\partial  \mathcal{E}}{\partial \theta^{k}_{\phantom{i}}}  &= \sum_i \frac{\partial  \mathcal{E}}{\partial \phi^{k+1}_i} \frac{\partial \phi^{k+1}_i }{\partial \theta^{k}_{\phantom{i}}}
,
\end{align}
where $k < K$, and we sum over all the intermediate indices of the derivatives.   The specific derivations will of course depend on the form of $f$, $g$ and $\mathcal{D}$, for which we give examples below.

\section{Applications to Markov random fields}
\label{sec:application-mrf}
It is easy to show that conventional sigmoid networks can be obtained by unfolding and untying mean field inference on binary pair-wise Markov random fields.  Although generic MRFs are not a good example of incorporating problem-level knowledge, it is instructive to consider very general graphical model formulation, both in order to understand conventional networks in terms of unfolding MRFs, and to generalize conventional deep networks by changing the model and/or the inference algorithm prior to unfolding.

Here we first review how mean-field updates can lead to conventional sigmoid networks.  Then we show how belief propagation leads to a different deep architecture.  Finally, we unify the two architectures using a general power mean formulation.

For simplicity we restrict discussion to pairwise MRFs. More general MRFs with higher-order factors can be easily expressed as pairwise MRFs by creating an auxiliary random variable for each higher-order factor. First we give a general formulation with arbitrary state spaces, and then discuss the special case binary MRFs which lead to sigmoid networks when unfolded.  Also for simplicity, we partition the variables into hidden and observed random variables, and omit connections between observed variables, since these do not affect inference.  

A pairwise MRF is represented here by an undirected graph whose vertices index hidden random variables $h_i$ taking values in  $\mathcal{H}_i$ for $i$ in $\mathcal{I}_{\mathsf{h}} = \{1, \ldots , N_{\mathsf{h}}\}$ and observed variables $ v_l$ taking values in $\mathcal{V}_{l}$ for $l$ in $\mathcal{I}_{\mathsf{v}} = \{1, \ldots , N_{\mathsf{v}}\}$ .  
We abuse notation by using $h_i$, $v_l$ to refer to both random variables and their values and by omitting their ranges in summations.  The factors of the probability distribution are associated with edges of the graph.  Edges between hidden variables are identified by unordered pairs of indices $(i,j)\equiv (j,i)$ in edge set $\mathcal{E}_{\mathsf{hh}} \defeq \{(i,j): i \text{ and } j \text{ are connected} \}$.   The set of edges $(i,l)$ between hidden and observed is $\mathcal{E}_{\mathsf{hv}} \defeq \{(i,l):  i \text{ and } l \text{ are connected} \}$.   The neighborhoods of node $i$ are $ \mathcal{N}^{\mathsf{hh}}_i \defeq \{j | (i,j) \in \mathcal{E}_{\mathsf{hh}}\}$, for hidden nodes, and $ \mathcal{N}^{\mathsf{hv}}_i = \{l | (i,l) \in \mathcal{E}_{\mathsf{hv}}\}$ for visible nodes.  The edge factors between hidden variables are parameterized by log potential functions  $\Psi(h_i,h_j) \defeq \Psi_{\mathsf{h}_i,\mathsf{h}_j}(h_i,h_j)$, and the hidden to visible potentials by $\Psi(h_i,v_l)\defeq \Psi_{\mathsf{h}_i,\mathsf{v}_l}(h_i,v_l)$,  where we again abuse notation by indexing the functions using their arguments.   

The MRF posterior probability distribution can then be written,
\begin{align}
p(h|v) 
&=\frac{1}{z(\Psi,v)}
	\prod_{(i,j) \in \mathcal{E}_{\mathsf{hh}}}
	e^{ \Psi(h_i,h_j)}  
	\prod_{(i,l) \in \mathcal{E}_{\mathsf{hv}}} 
	e^{\Psi(h_i,v_l)},\\
&\propto
	\exp\Bigl({\displaystyle \sum_{(i,j) \in \mathcal{E}_{\mathsf{hh}}}
	\Psi(h_i,h_j) + \sum_{(i,l) \in \mathcal{E}_{\mathsf{hv}}} 
	\Psi(h_i,v_l)}\Bigr).
	\label{eq:mrf_model_hv}
\end{align} 
For discrete $h_i$ and $v_l$ the log potential functions are typically represented using scalar parameters for each combination of values taken by their arguments, and the MRF can be formulated as an exponential family model using indicator functions as features.

\subsection{Mean field inference}
In variational methods, of which the \emph{mean field} (MF) approximation is a special case, we perform a tractable approximate inference by minimizing Kullback-Leibler (KL) divergence between an approximate posterior $q_{\mathsf{h}}$ and the true posterior $p_{\mathsf{h} \mid \mathsf{v}}$.  Equivalently we maximize a lower bound on the likelihood obtained via Jensen's inequality:
\begin{align}
 \arg\min_{q_{\mathsf{h}}} &D_{KL}(q_{\mathsf{h}} || p_{\mathsf{h}|\mathsf{v}}) 
 =  \arg\max_{q_{\mathsf{h}}} \mathcal{L}(q_{\mathsf{h}}, p_{\mathsf{h},\mathsf{v}}), 
\\
\mathcal{L}(q_{\mathsf{h}}, p_{\mathsf{h},\mathsf{v}}) & \stackrel{\mathrm{def}}{=} \sum_{h} q(h) \log \frac{p(h,v)}{q(h)} \leq \log p(v).
\end{align}
In the mean field approximation, the posterior is fully factorized over the variables so that  $q(h) = \prod_{i \in \mathcal{I}_{\mathsf{h}}} q(h_i)$, which is a product of marginal posteriors.  
This leads to bound-preserving update equations of the form,
\begin{align}
q(h_i) \propto & 
	\exp\Bigl( 
		\sum_{j \in \mathcal{N}^{\mathsf{hh}}_i}
		\sum_{h_j} 
		q(h_j) \Psi(h_i,h_j)
		+ \sum_{l \in \mathcal{N}^{\mathsf{hv}}_i}
		\Psi(h_i,v_l^{\ast}) 
	\Bigr),
\label{eq:mfupdate}		
\end{align}
where $\sum_{h_i} q(h_i) = 1$, and $v_l^{\ast}$ is the observed value of $\mathsf{v}_j$.  
Normalizing $q(h_i)$ leads to a multivariate logistic or ``sigmoid'' function,
\begin{align}
q(h_i) = & 
	\frac{	\exp\Bigl( 
			\sum_{j \in \mathcal{N}^{\mathsf{hh}}_i}
			\sum_{h_j} 
			q(h_j) \Psi(h_i,h_j)
			+ \sum_{l \in \mathcal{N}^{\mathsf{hv}}_i}
			\Psi(h_i,v_l^{\ast}) 
		\Bigr)
	}
	{\sum_{h_i'} \exp\Bigl( 
		\sum_{j \in \mathcal{N}^{\mathsf{hh}}_i}
		\sum_{h_j} 
		q(h_j) \Psi(h_i',h_j)
		+ \sum_{l \in \mathcal{N}^{\mathsf{hv}}_i}
		\Psi(h_i',v_l^{\ast}) 
		\Bigr)
	}.
\label{eq:mfupdate}		
\end{align}
Here we formulate the updates in terms of messages to facilitate later comparison with belief propagation:  
\begin{align}
q(h_i) \propto & \exp\Bigl( 
		\sum_{j \in \mathcal{N}^{\mathsf{hh}}_i}
		\log \msg{j}{i}(h_i)
		+ \sum_{l \in \mathcal{N}^{\mathsf{hv}}_i}
		\Psi(h_i,v_l^{\ast})
		\Bigr),
\label{eq:mfbelief}		
\end{align}
where messages $\msg{j}{i}(h_i)$ from $j$ to $i$ at value $h_i$ are given by
\begin{align}
\msg{j}{i}(h_i) & \propto \exp\Bigl(\sum_{h_j  } q(h_j) \Psi(h_i,h_j)\Bigr).
\label{eq:mfmsg}
\end{align}
Note that messages in \ref{eq:mfbelief} can be unnormalized, or  normalized as needed for numerical purposes, however $q(h_j)$ used in \eqref{eq:mfupdate} and \eqref{eq:mfmsg} must be normalized.
In order to maintain the variational bound, the updates must be done according to an update schedule that avoids synchronous updates of directly interdependent $q$ functions.  However, in the context of discriminative training, with an unfolded model, maintaining the bound may not be necessary.  Nevertheless, the specific ordering of updates may have a strong effect on the rate of convergence of inference.    We can implement an arbitrary schedule using a scalar $\msgax{\alpha}{j}{i}{k}$ in each iteration $k$, so that the message is computed if $\msgax{\alpha}{j}{i}{k} = 1$ and the previous value of the message is kept if $\msgax{\alpha}{j}{i}{k} = 0$.   Unfolding the algorithm, and untying the parameters can then be formulated as 
\begin{align}
q^{k}(h_i) 
\propto & 
	\exp\Bigl(
	\sum_{j \in \mathcal{N}^{\mathsf{hh}}_i}
	\log \msgax{\tilde{m}}{j}{i}{k}(h_i)
	+ \sum_{l \in \mathcal{N}^{\mathsf{hv}}_i}
	\Psi^{k}(h_i,v_l^{\ast})\Bigr),
\label{eq:mfbelief_unfold}		
\end{align}
with the schedule implemented by a power mean of the previous and new message, here parameterized using exponent $\rho$ to allow us to choose the type of mean used, including, for example, arithmetic, with $\rho = 1$,  or geometric, with $\rho \rightarrow 0$ (see Appendix \ref{sec:mean} for a review of power means).  
\begin{align}
\msgax{\tilde{m}}{j}{i}{k}(h_i) \propto \Bigl(\msgax{\alpha}{j}{i}{k} \msgax{m}{j}{i}{(k)\rho}(h_i)
+ (1 - \msgax{\alpha}{j}{i}{k}) \msgax{\tilde{m}}{j}{i}{(k-1)\rho}(h_i)\Bigr)^{\frac{1}{\rho}}, 
\label{eq:mfmsg_alpha}
\end{align}
with messages
\begin{align}
\msgax{m}{j}{i}{k}(h_i) & \propto \exp\Bigl(\sum_{h_j  } q^{k-1}(h_j) \Psi^{k}(h_i,h_j)\Bigr).
\label{eq:mfmsg_unfold}
\end{align}
In this way we can consider optimizing the schedule, if desired, by allowing $\msgax{\alpha}{j}{i}{k} \in [0,1]$, or for example, perform synchronous updates by setting $\msgax{\alpha}{j}{i}{k} = 1$ for all $i,j$.  This scheduling formulation can be used with any message-passing inference algorithm, but different choices of arithmetic versus geometric mean might be more convenient with different messages.  For mean field messages, the geometric mean seems more convenient, given the log-linear form of \eqref{eq:mfmsg_unfold}.   Under this choice we can take $\alpha$ inside the message to yield,
\begin{align}
\msgax{m}{j}{i}{k}(h_i) & \propto \exp\Bigl(\msgax{\alpha}{j}{i}{k} \sum_{h_j  } q^{k-1}(h_j) \Psi^{k}(h_i,h_j) + (1 - \msgax{\alpha}{j}{i}{k}) \log \msgax{m}{j}{i}{k-1}(h_i) \Bigr).
\label{eq:mfmsg_unfold}
\end{align}

To map this back to the general formulation, we can define, for example, 
$$ \phi^{k} = q^{k}, \quad \theta^{k} = \{\Psi_{\mathsf{hh}}^{k},\Psi_{\mathsf{hv}}^{k},\alpha^{k}\}, \quad  y^{K} = g(q^K),$$ for some estimator $g$.    

It is interesting to draw a comparison with conventional sigmoid neural networks.
To do so, we consider an MRF with binary random variables, $h_i$, $v_l\in \{0,1\} = \mathcal{H}_i = \mathcal{V}_l$.   The MRF posterior distribution can be written, 
\begin{align}
p(h|v) \propto 
\exp\Bigl(
	\sum_{i,j \in \mathcal{I}_{\mathsf{h}}} 
		{\textstyle \frac{1}{2}} a_{i,j} h_i  h_j 
	+  \sum_{i \in \mathcal{I}_{\mathsf{h}}} 
		b_i h_i 
	+ \sum_{{i \in \mathcal{I}_{\mathsf{h}}, l \in \mathcal{I}_{\mathsf{v}}}} 
 	    c_{i,l}  h_i v_l^{\ast}  
\Bigr),
\end{align}
with suitable choices for $a_{i,j}$, $b_{i}$, and $c_{i,l}$, derived from 
$\Psi$ (see Appendix \ref{sec:binaryMRF}).  The factor of $1/2$ comes from the fact that $a_{i,j} = a_{j,i}$, and each edge potential is counted twice in the sum.   In matrix notation, with $A = \{a_{i,j}\}_{i,j \in \mathcal{I}_{\mathsf{h}}}$, where $a_{i,j}=0$ for $(i,j) \notin \mathcal{E}_{\mathsf{hh}}$, and similarly for matrix $C$, and vector $b = \{b_{i}\}_{i\in \mathcal{I}_{\mathsf{h}}}$.  We can then write the desired posterior as: 
\begin{align}
p(h|v) \propto \exp\left({ {\textstyle \frac{1}{2}} h^T A h  + h^T b  + h^T C v }\right)
\label{eq:binary-mrf}
\end{align}
Note that $a_{i,i}=0$ in the original model since there are no self-edges.  Unfolding and untying parameters as in \eqref{eq:mfbelief_unfold}, and using synchronous updates with $\alpha = 1$ in \eqref{eq:mfmsg_alpha}, then leads to 
\begin{align}
\mu^{k} = \operatorname{logistic}\left({A}^{k} \mu^{k-1} + b^{k}  + C^{k}v^{\ast}\right)
\label{eq:sigmoidnet}
\end{align}
where $\mu^{k} \defeq \{\mu_i^k\}_{i \in \mathcal{I}_{\mathsf{h}}}$, and $\mu_i^{k} \defeq q^{k}(h_i = 1)$. This can be recognized as a sigmoid network having a special structure in which inputs are connected to all the layers.  This structure is a consequence of unfolding a model in which any hidden variables may be directly connected to observations.   However, as we can untie the parameters in any way we please, to emulate the conventional case where the first layer depends only on the inputs and each subsequent layer only depends on the previous one, we can allow $c_{i,l}^{k}$ to be non-zero only on the first frame, $k = 0$.  The initial distribution $\mu_i^{k=0}$, as well as the associated weights,  $a_{i,j}^{k=1}$,  can be set to zero. We can also relax the constraint that $a_{i,i}^{k} = 0$ from the original model to reach the full generality of the conventional sigmoid network.

It is worth noting that conventional feed-forward sigmoid networks can also be derived, more simply, by starting with a deep, layer-wise binary MRF, and performing a single forward pass of mean-field updates starting with the input and ending with the last layer.  This corresponds to a special case of MRF structure and update schedule in our framework.  Our point in the above is to illustrate that the more general case of MRF of arbitrary structure can also lead to a feed-forward sigmoid network.  When looking at a given conventional neural network, then, we may be able to interpret it in two different ways. We can either interpret it as an approximate MF inference in an MRF with the same structure as the neural network, or in some cases we may be able to interpret it as a deeper unfolding of a model with a more compact structure.  In either case, once one has identified a model and inference algorithm that unfold into a given neural network, one can consider changing the inference algorithm or model structure in order to generate alternative variants of the neural network.   For example, instead of using mean-field inference, one could unfold the model using belief propagation.  

\subsection{Belief propagation}
\emph{Belief propagation} (BP) is an algorithm for computing posterior probabilities, which leads to an exact solution for tree-structured graphical models \cite{pearl1988probabilistic}.  When applied to graphs with loops it is known as \emph{loopy belief propagation}. It can be interpreted as a fixed point algorithm  for the stationary points of the Bethe free energy \cite{yedidia2005constructing}, which in turn can be seen as an approximation to the Kullback-Leibler divergence between the approximated posterior and the true posterior distribution.    Algorithms in the style of belief propagation have been thought to produce better results on general Markov random field problems \cite{weiss2001comparing}, and hence there is a motivation to investigate deep network architectures based on BP.  
Some previous work has explored unfolding of BP without untying the parameters \cite{domke2011parameter,domke2013learning}, which focused on an extension to loopy BP based on TRW-BP approaches \cite{wainwright2005new}, but for simplicity we begin with the standard sum-product version of BP.  

In BP as in MF methods, the update equations are formulated in terms of marginal posteriors, known as beliefs, based on messages: 
\begin{align}
q(h_i) 
&\propto
	\prod_{j \in \mathcal{N}^{\mathsf{hh}}_i} 
		\msg{j}{i} (h_i) 
		\prod_{l \in \mathcal{N}^{\mathsf{hv}}_i} 
		e^{\Psi(h_i,v_l^{\ast})}, 	
\label{eq:bpupdate}
\end{align}    
where $\sum_{h_i} q(h_i) = 1$, with messages defined by  
\begin{align}
\msg{j}{i}(h_i) 
\propto \sum_{h_j}  	
	\frac{q(h_j)}{ \msg{i}{j}(h_j)}
	e^{\Psi(h_i,h_j)}.
\label{eq:bpmsg}	 
\end{align}
As in the MF updates, normalization of messages is optional.  However, in contrast to MF, in which the beliefs have to be normalized in each iteration, the normalization of beliefs in BP is optional and can be done whenever desired for numerical reasons, or to compute output predictions.
For comparison with MF equations, we formulate \eqref{eq:bpupdate} as:
\begin{align}
q(h_i) \propto & \exp\Bigl( 
		\sum_{j \in \mathcal{N}^{\mathsf{hh}}_i}
		\log \msg{j}{i}(h_i)
		+ \sum_{l \in \mathcal{N}^{\mathsf{hv}}_i}
		\Psi(h_i,v_l^{\ast})
		\Bigr),
\label{eq:bpbelief}
\end{align}
and see that \eqref{eq:bpbelief}, is identical to \eqref{eq:mfbelief}, so that only the messages differ between MF \eqref{eq:mfmsg} versus BP \eqref{eq:bpmsg}.

For tree-structured graphs, the exclusion of the incoming message $\msg{i}{j}(h_j)$ in \eqref{eq:bpmsg} prevents ``feedback'' by ensuring that each message is only incorporated once into a given belief, and the updates yield exact marginals, from which the full posterior can be computed.  
In the general case where MRFs may have cycles, the exclusion of incoming messages no longer completely prevents feedback, and the approximate marginals no longer are guaranteed to converge to the true marginals.  However loopy BP works well in practice for some problems, with an appropriate message-passing schedule.  

As in the mean-field updates, we can formalize a message-passing schedule by parameterizing the update in \eqref{eq:bpmsg} with a scalar $\msgax{\alpha}{j}{i}{k}$ in each iteration $k$.  
Since the two algorithms only differ in terms of the messages, unfolding and untying the algorithm across layers, and implementing the update schedule,  is accomplished using \eqref{eq:mfbelief_unfold} and \eqref{eq:mfmsg_alpha} with the unfolded BP messages,
\begin{align}
\msgax{m}{j}{i}{k}(h_i)
&\propto 
\sum_{h_j}  	
	\frac{q^{k-1}(h_j)}{{\msgax{m}{i}{j}{k-1} (h_j)}}	
	e^{\Psi^{k-1}(h_i,h_j)}.
\label{eq:bpmsg_unfold}
\end{align}
We can then consider optimizing the message passing schedule as part of the deep unfolding method rather than using heuristics as is commonly done.  Since the BP messages \eqref{eq:bpmsg_unfold} are linear instead of log-linear, a choice of arithmetic mean ($\rho = 1$) in \eqref{eq:mfmsg_alpha} seems more convenient.  This leads to messages of the form, 
\begin{align}
\msgax{m}{j}{i}{k}(h_i)
&\propto 
\msgax{\alpha}{j}{i}{k} 
\sum_{h_j}  	
	\frac{q^{k-1}(h_j)}{{\msgax{m}{i}{j}{k-1} (h_j)}}	
	e^{\Psi^{k-1}(h_i,h_j)}
+ 
(1 - \msgax{\alpha}{j}{i}{k})
\msgax{m}{j}{i}{k-1}(h_i)
.
\label{eq:bpmsg_unfold}
\end{align}

As in the MF case, we can obtain similar updates by starting with a layer-wise graph structure, with an update schedule that passes sequentially through the layers, in the manner of a feed-forward neural network.   
However as in the MF case, the general unfolding framework allows us to consider more structured base models and still obtain a feedforward architecture.  In general when untying and training discriminatively we can also potentially neglect the  elimination of the incoming message, expressed by  $\msg{i}{j}(h_j)$  in \eqref{eq:bpmsg_unfold}, since the untied parameters imply that the incoming nodes can be considered as different random variables.  

\subsection{Generalized message passing}
We now consider unifying the belief propagation messages with the mean-field messages in order to develop an architecture that encompasses both as special cases. 
Leaving aside the update schedule, comparing \eqref{eq:mfmsg} with \eqref{eq:bpmsg}	 
we can see that
the two functions are related via a general weighted power mean $M_{a}(w,x) =  \left(\sum_n w_n x_n^a \right)^{1/a}$ (see Appendix \ref{sec:mean}), where $\sum_n w_n = 1$, with the parameters defined as
\begin{align}
w_{n} = q^{k-1}(h_j=n)
,\quad 
x_{n} = \frac{e^{\Psi^{k}(h_i,h_j=n)}}{{\msg{i}{j}(h_j=n)}}
,\quad  
a = \lambda_{i,j}^{k}, 
\end{align}
\begin{align}
\msgax{\tilde{m}}{j}{i}{k}(h_i) \propto 
	\Bigl(
		\sum_{h_j} 
		\frac{q^{k-1}(h_j)}{(\msgax{m}{i}{j}{k-1} (h_j))^{\lambda_{i,j}^{k}}}
		e^{\lambda_{i,j}^{k} \Psi^{k}(h_i,h_j)}
	\Bigr)^{\frac{1}{\lambda_{i,j}^{k}}}	
\label{eq:unified_msg}
\end{align}
so that $\lambda_{i,j}^{k} = 1$ exactly yields BP sum-product messages. The limit $\lambda_{i,j}^{k} \rightarrow 0$ yields the MF updates:   
\begin{align}
\lim_{\lambda_{i,j}^{k} \rightarrow 0} 
\msgax{\tilde{m}}{j}{i}{k}(h_i)
& \propto
	\exp\Bigl(
		\sum_{h_j} 
		q^{k-1}(h_j)
		\Bigl[
		\Psi^{k}(h_i,h_j)	
		- \log \msgax{m}{j}{i}{k-1}(h_j)
		\Bigr]				
	\Bigr)
\nonumber \\	
& \propto
	\exp\Bigl(
		\sum_{h_j} 
		q^{k-1}(h_j)
		\Psi^{k}(h_i,h_j)
	\Bigr),  
\label{eq:unified_msg_mf}
\end{align}
since $c =  - \sum_{h_j} q^{k-1}(h_j)  \log \msgax{m}{j}{i}{k-1}(h_j)$ is constant in terms of $h_i$.

It is also possible to interpolate between sum-product and max-product varieties of BP by applying the power mean differently, 
and that can be done in combination with \eqref{eq:unified_msg}, at the cost of clarity, to yield an even more flexible general form (see Appendix \ref{sec:generalized_forms}).
It is an interesting empirical question to what extent these distinctions affect performance in the context of deep unfolding.

Other works \cite{hazan2010norm, wiegerinck2003fractional} have generalized between different forms of BP.  A generalization similar to ours between BP, TRW-BP, and MF is described in \cite{liu2012negative}, in which negative $\lambda_{i,j}$ are also considered.   
It is interesting to note that TRW-BP \cite{wainwright2005new}, and related BP algorithms, maintain a bound on the inference time objective function using edge weights similar to $\lambda$, as well as specific message-passing schedules, as here implemented by $\alpha$ using \eqref{eq:unified_msg} in \eqref{eq:mfmsg_alpha}.  However, in the context of training a deep network, maintaining a bound may not be nearly as important as the general form of the activation function.  This is because rather than optimizing an objective function at test time, we instead optimize the parameters according to a discriminative objective function at training time, such that the inference algorithm implemented by the network obtains the best result on training data.

Our generalized form raises the possibility of starting with a trained conventional sigmoid network, and re-training it to perform BP updates, by incrementally changing a globally tied $\lambda$ from $0$ to $1$, in order to investigate differences in performance between the two architectures.   In this case, in \eqref{eq:unified_msg}, we must gracefully handle the numerical instability when $\lambda\rightarrow 0$ by, for example, doing an interpolation near $0$.  
To use an existing sigmoid network defined according to \eqref{eq:sigmoidnet} with the generalized update \eqref{eq:unified_msg}, we can easily derive $\Psi$ from $A$ and $b$, as shown in Appendix \ref{sec:binaryMRF}.

The optimization of parameters via the chain rule is relatively straightforward, and the above unfolded algorithms can be optimized by supplying a task-related objective function and computing derivatives of the message passing formulae above.   

When optimizing the schedule, $\alpha$ should remain in $[0,1]$, which can be accomplished, for example, by parameterizing them as  $\alpha = 1/(1 + \exp(-z))$, with $z \in \mathbb{R}$.   
We can also consider training $\lambda$ via back-propagation.  
Clearly the schedule parameters $\alpha$ and the message style parameters $\lambda$ can add a significant number of parameters.   Assuming uniform state spaces for all variables, $\alpha$ and $\lambda$ are both of size $|\mathcal{I}_{\mathsf{h}}|^2$ whereas in general $\Psi$ could have a maximum size of $ |\mathcal{I}_{\mathsf{h}}|^2|\mathcal{H}|^2 +|\mathcal{I}_{\mathsf{h}}||\mathcal{I}_{\mathsf{v}}||\mathcal{H}||\mathcal{V}| $.  
As a consequence, judicious tying may be necessary to keep complexity in check.  For example, one $\lambda^{k}$ for each layer might be a reasonable level of tying depending on the network architecture.  Training individual $\lambda_{i,j}^{k}$ for each connection might be excessive if there are few states per node, such as with binary units.   However for multinomial networks with many states ($|\mathcal{H}_i| \gg 2$), tuning $\lambda_{*,j}^{k}$ (shared across inputs for the same output unit) might be more reasonable.

One caveat in the optimization is that sigmoid networks are known to be difficult to train relative to simpler activation functions such as \emph{maxout} \cite{goodfellow2013maxout}, due to the vanishing of gradients.
The normalized BP activation function and its generalizations, especially in the binary form shown in Appendix \ref{sec:binaryMRF}, may be even more complicated than the corresponding sigmoid activation.  
However, the unnormalized log form, using the generalized BP equations \eqref{eq:generalized-sumprod-maxprod} and \eqref{eq:bpbelief}, neglecting the schedule, and considering the reciprocal messages $\msgax{m}{i}{j}{k-1}(h_j)$ to be uniform, leads to the updates,
\begin{align}
u^{k}(h_i) 
=& 
		\sum_{j \in \mathcal{N}^{\mathsf{hh}}_i}
		{\frac{1}{\kappa}}	
		\log 
		\Bigl(
			\sum_{h_j} 
			\frac{1}{N_{\mathsf{h_j}}}			
			\exp
			\bigl(
			    \kappa u^{k-1}(h_j) + \kappa \Psi^{k}(h_i,h_j)
			\bigr)
		\Bigr)
		+ \sum_{l \in \mathcal{N}^{\mathsf{hv}}_i}
		\Psi^{k}(h_i,v_l^{\ast})
\label{eq:logsumprod}
\\
\underset{\kappa \rightarrow \infty}{=} & 
		\sum_{j \in \mathcal{N}^{\mathsf{hh}}_i}
			\max_{h_j} 			
			\bigl(
			    u^{k-1}(h_j) 
			    + 
			    \Psi^{k}(h_i,h_j)
			\bigr)
		+ \sum_{l \in \mathcal{N}^{\mathsf{hv}}_i}
		\Psi^{k}(h_i,v_l^{\ast}).
\label{eq:logmaxprod}
\end{align}
This appears comparable to two commonly used forms: \eqref{eq:logsumprod} is similar in spirit to  \emph{softmaxout} \cite{zhang2014improving}, and the max-product version \eqref{eq:logmaxprod} is similar in spirit to maxout \cite{goodfellow2013maxout}.   Despite the similarity, the methods are not the same, and it remains to be seen whether the differences are significant in practice.   We leave experiments on generic MRFs for other work, and in the rest of this paper we turn to models that incorporate specific problem domain knowledge.

\section{Deep non-negative matrix factorization}
\label{sec:app:nmf}
While discrete MRFs are an interesting general case, one of the main points of this work is to incorporate problem-level knowledge into a novel deep architecture.  To that end, here we apply the proposed deep unfolding framework to the non-negative matrix factorization (NMF) model \cite{Lee01-AFN}, which can be applied to any non-negative signal.
Although NMF can be applied in many domains, here we focus on the task of single-channel source separation, which aims to recover source signals from mixtures.
In this context it encompasses the simple problem-level assumptions that power or magnitude spectra of different sources approximately add together, and that each source can be described as a linear combination of non-negative basis functions.

NMF operates on a matrix of $F$-dimensional non-negative spectral features, usually the power or magnitude spectrogram of the mixture,
${\bf M} = [ {\bf m}_1 \cdots {\bf m}_T ]$,
where $T$ is the number of frames and ${\bf m}_t \in \setR_+^F$, $t=1, \dots, T$ are obtained by short-time Fourier transformation of the time-domain signal.
With $L$ sources, each source $l \in \{1, \dots, L\}$ is represented using a matrix containing $R_l$ non-negative basis column vectors, $ {\bf W}^l = \{{\bf w}^l_r\}_{r=1}^{R_l} $, multiplied by a matrix of activation column vectors ${\bf H}^l = \{{\bf h}^l_t\}_{t=1}^{T}$, for each time $t$.  The $r$th row of  ${\bf H}^l$ contains the activations for the corresponding basis ${\bf w}^l_r$ at each time $t$.
A column-wise normalized $\widetilde{\bW}^l$ can be used to avoid scaling indeterminacy.  
The basic assumptions can then be written as
\begin{equation}
{\bf M} \approx \sum_l {\bf S}^l \approx \sum_l \widetilde{\bW}^l{\bH}^l = \widetilde{\bW}{\bH} .
\label{eq:nmf}
\end{equation}
The $\beta$-divergence, $D_{\beta}$, is an appropriate cost function for this approximation \cite{Fevotte09-NMF},
which casts inference as an optimization of $\hat{\bf H},$
\begin{equation}
\hat{\bf H} =
\argmin\limits_{\bf H} D_{\beta}({\bf M} \mid \widetilde{\bW} {\bf H}) + \mu|{\bf H}|_1.
\label{eq:sup_nmf_obj}
\end{equation}
For $\beta=1$, $D_{\beta}$ is the generalized KL divergence, whereas  $\beta = 2$ yields the squared error. An L1 sparsity constraint with weight $\mu$ favors solutions where few basis vectors are active at a time.

The following multiplicative updates minimize \eqref{eq:sup_nmf_obj} subject to non-negativity constraints \cite{Fevotte09-NMF},
\begin{align}
{\bf H}^{k} = {\bf H}^{k-1} \circ \frac{
    \displaystyle \widetilde{\bW}^{T} ( {\bf M} \circ (\widetilde{\bW} {\bf H}^{k-1})^{\beta-2} )
}{
    \displaystyle \widetilde{\bW}^{T} (\widetilde{\bW} {\bf H}^{k-1})^{\beta-1} + \mu
},
\label{eq:nmf_updates}
\end{align}
for iteration $k \in \{1, \ldots, K\}$, where $\circ$ denotes element-wise multiplication, the matrix quotient is element-wise, and ${\bf H}^0$ is initialized randomly.

After $K$ iterations, to reconstruct each source, typically a Wiener filtering-like approach is used, which enforces the constraint that all the source estimates $\widetilde{\bf S}^{l,K}$ sum up to the mixture:
\begin{equation}
\widetilde{\bf S}^{l,K} = \frac{\widetilde{\bW}^{l} {\bf H}^{l,K}}{\sum_{l'} \widetilde{\bW}^{l'} {\bf H}^{l',K}} \circ {\bf M}.
\label{eq:wiener}
\end{equation}

While in general, NMF bases are trained independently on each source before being combined, the combination is not trained discriminatively for good separation performance from a mixture. Recently, discriminative methods have been applied to sparse dictionary based methods to achieve better performance in particular tasks \cite{mairal2012task}. In a similar way, we can discriminatively train NMF bases for source separation. The following optimization problem for training bases, termed \emph{discriminative NMF} (DNMF) was proposed in \cite{DNMF,sprechmann2014hscma}:
\vspace{-0.1cm}
\begin{align}
&\hat{\bf W} = \argmin\limits_{\bf W}
\sum_l  \gamma_l D_{\beta_{2}}\left({\bf S}^l \mid {{\bf W}}^l \hat{\bf H}^l({\bf M},{\bf W})\right),
\label{eq:real_dnmf} \\
&  \hat{\bf H}({\bf M},{\bf W}) =
\argmin\limits_{{\bf H}} D_{\beta_{1}}({\bf M} \mid \widetilde{{\bf W}} {\bf H}) + \mu|{\bf H}|_1,
\label{eq:real_dnmf_h}
\end{align}
and where $\beta_{1}$ controls the divergence used in the bottom-level analysis objective, and $\beta_{2}$ controls the divergence used in the top-level reconstruction objective. The weights $\gamma_l$ account for the application-dependent importance of source $l$; for example, in speech de-noising, we focus on reconstructing the speech signal.  The first part \eqref{eq:real_dnmf} minimizes the reconstruction error given $\hat{\bf H}$.   The second part ensures that $\hat{\bf H}$ are the activations that arise from the test-time inference objective.
Given the bases ${\bf W}$, the activations $\hat{\bf H}({\bf M},{\bf W})$ are uniquely determined, due to the convexity of  \eqref{eq:real_dnmf_h}. Nonetheless, the above remains a difficult  bi-level optimization problem, since the bases ${\bf W}$ occur in both levels.
 	
In \cite{sprechmann2014hscma} the bi-level problem was approached by directly solving for the derivatives of the lower level problem after convergence.
In \cite{DNMF}, the problem was approached by untying the bases used for reconstruction in \eqref{eq:real_dnmf} from the analysis bases used in \eqref{eq:real_dnmf_h}, and training only the reconstruction bases.  In addition, \eqref{eq:wiener} was incorporated into the discriminative criteria as
\begin{align}
&\hat{\bf W} = \argmin\limits_{\bf W}
\sum_l  \gamma_l D_{\beta_{2}}\left({\bf S}^l \mid \widetilde{\bf S}^{l,K}({\bf M},{\bf W}) \right),
\label{eq:real_dnmf_wiener}
\end{align}
This model can be considered a first step toward the proposed approach in the context of NMF.

Here, based on our framework, we unfold the entire model as a deep non-negative neural network, and we untie the parameters across layers as $\bW^{k}$ for $k =1, \ldots, K$. We call this new model \emph{deep NMF}.
We cast this into our general formulation by defining
$$i=t, \quad x_i = {\bf m}_t, \quad y^{*} = {\bf S}^{l}, \quad \phi^{k} = \bH^{k}, \quad y^{K} = \widetilde{\bf S}^{l,K}, \quad \theta^{k} = \bW^{k}.$$
We identify the inference objective and estimator \eqref{eq:general:obj1} with \eqref{eq:real_dnmf_h} and \eqref{eq:wiener},  the discriminative objective \eqref{eq:general:obj2} with \eqref{eq:real_dnmf_wiener}, and the iterative updates \eqref{eq:general:update1} with \eqref{eq:nmf_updates}.

In order to train this network while respecting the non-negativity constraints, we derive recursively defined multiplicative update equations by back-propagating a split between positive and negative parts of the gradient. 
In NMF, multiplicative updates are often derived using a heuristic approach which uses the ratio of the negative part to the positive part as a multiplication factor to update the value of that variable of interest.   Here we do the same for each $\bW^{k}$ matrix in the unfolded network:
\begin{align}
\bW^{k} \Leftarrow \bW^{k} \circ \frac{\left[ \nabla_{\bW^{k}} \mathcal{E} \right]_-} {\left[ {\nabla_{\bW^{k}} \mathcal{E}} \right]_+}.
\end{align}
To propagate the positive and negative parts, we use:
\begin{align}
\left[
	\frac{\partial \mathcal{E}}
	{\partial h^{k}_{r_k,t}}
\right]_+ &= 
\sum_{r_{k+1}} 
\left( 
	\left[
		\frac{\partial \mathcal{E}}
		{\partial h^{k+1}_{r_{k+1},t}} 
	\right]_+
	\left[
		\frac{\partial h^{k+1}_{r_{k+1},t} }
		{\partial h^{k}_{r_{k},t}} 
	\right]_+
	+	
	\left[
		\frac{\partial \mathcal{E}}
		{\partial h^{k+1}_{r_{k+1},t}} 
	\right]_-
	\left[
		\frac{\partial h^{k+1}_{r_{k+1},t} }
		{\partial h^{k}_{r_k,t}}
	\right]_- 
\right), \\
\left[
	\frac{\partial \mathcal{E}}
	{\partial h^{k}_{r_k,t}}
\right]_- &= 
\sum_{r_{k+1}} 
\left( 
	\left[
		\frac{\partial \mathcal{E}}
		{\partial h^{k+1}_{r_{k+1},t}} 
	\right]_+
	\left[
		\frac{\partial h^{k+1}_{r_{k+1},t} }
		{\partial h^{k}_{r_{k},t}} 
	\right]_-
	+	
	\left[
		\frac{\partial \mathcal{E}}
		{\partial h^{k+1}_{r_{k+1},t}} 
	\right]_-
	\left[
		\frac{\partial h^{k+1}_{r_{k+1},t} }
		{\partial h^{k}_{r_k,t}}
	\right]_+ 
\right), \\
\left[
	\frac{\partial \mathcal{E}}
	{\partial w^{k}_{f,r}} 
\right]_+ &=
\sum_t 
\sum_{r_{k+1}} 
\Bigg( 
	\left[ 
		\frac{\partial \mathcal{E}}
		{\partial h^{k+1}_{r_{k+1},t}}
	\right]_+ 
	\left[
		\frac{\partial h^{k+1}_{r_{k+1},t} }
		{\partial w^{k}_{f,r}}
	\right]_+ 
	+ 
	\left[ 
		\frac{\partial \mathcal{E}}
		{\partial h^{k+1}_{r_{k+1},t}}
	\right]_- 
	\left[
		\frac{\partial h^{k+1}_{r_{k+1},t} }
		{\partial w^{k}_{f,r}}
	\right]_- 
\Bigg),\\
\left[
	\frac{\partial \mathcal{E}}
	{\partial w^{k}_{f,r}} 
\right]_- &=
\sum_t 
\sum_{r_{k+1}} 
\Bigg( 
	\left[ 
		\frac{\partial \mathcal{E}}
		{\partial h^{k+1}_{r_{k+1},t}}
	\right]_+ 
	\left[
		\frac{\partial h^{k+1}_{r_{k+1},t} }
		{\partial w^{k}_{f,r}}
	\right]_- 
	+ 
	\left[ 
		\frac{\partial \mathcal{E}}
		{\partial h^{k+1}_{r_{k+1},t}}
	\right]_- 
	\left[
		\frac{\partial h^{k+1}_{r_{k+1},t} }
		{\partial w^{k}_{f,r}}
	\right]_+ 
\Bigg),
\end{align}
where $h^{k}_{r_k,t}$ are the activation coefficients at time $t$ for the $r_k$th basis set in the $k$th layer, and $w^{k}_{f,r}$ are the values of the $r$th basis vector in the $f$th feature dimension in the $k$th layer.
\section{Experiments}
\label{sec:experiments}

The deep NMF method was evaluated along with competitive models on the 2nd CHiME Speech Separation and Recognition Challenge corpus \footnote{http://spandh.dcs.shef.ac.uk/chime\_challenge/ -- as of June.~2014}.
The task is speech enhancement in reverberated noisy mixtures ($S=2$, $l=1$: speech, $l=2$: noise). The background is mostly non-stationary noise sources such as children, household appliances, television, radio, and so on, recorded in a home environment.   
Training, development, and test sets of noisy mixtures along with noise-free reference signals are created from the Wall Street Journal (WSJ-0) corpus of read speech and a corpus of training noise recordings. The dry speech recordings are convolved with time-varying room impulse responses estimated from the same environment as the noise.  The training set consists of 7\,138 utterances at six SNRs from -6 to 9~dB, in steps of 3~dB. The development and test sets consist of 410 and 330 utterances at each of these SNRs, for a total of 2\,460 / 1\,980 utterances.
By construction of the WSJ-0 corpus, our evaluation is speaker-independent. The background noise recordings in the development and test set are different from the training noise recordings, and different room impulse responses are used to convolve the dry utterances. In this paper, we present results on the development set.
To reduce complexity we use only 10\,\% of the training utterances for all methods. 
Our evaluation measure is source-to-distortion ratio (SDR) \cite{Vincent06-PMI}.

\subsection{Feature extraction}

Each feature vector
concatenates $T=9$ consecutive frames of left context, ending with the target frame,
obtained as short-time Fourier spectral magnitudes, using 25\,ms window size, 10\,ms window shift, and the square root of the Hann window. This leads to feature vectors of size $TF$ where $F=200$ is the number of frequencies.
Similarly to the features in ${\bf M}$, each column of $\hat{\bf S}^l$ corresponds to a sliding window of consecutive reconstructed frames.
Only the last frame in each sliding window is reconstructed, which leads to an on-line algorithm.
For the NMF-based approaches, we use the same number of basis vectors for speech and noise ($R^1 = R^2$), and consider $R^l=100$ and $R^l=1000$. We denote the total as $R = \sum_l R^l.$
We look at two regimes of maximum iterations, $K = 4$ for which NMF-based approaches still have significant room for improvement in performance, and $K=25$ for which, based on preliminary experiments, they are close to asymptotic performance.

\subsection{Baseline 1: Deep Neural Network}

To compare our deep NMF architecture with standard $K$-layer deep neural networks, we used the following setting.
The feed-forward DNNs have $K-1$ hidden layers with hyperbolic tangent activation functions and an output layer with logistic activation functions.
Denoting the output layer activations for time index $t$ by ${\bf y}_t = (y_{1,t}, \dots, y_{F,t})^T \in [0,1]^F,$
the DNN computes the deterministic function
$${\bf y}_t = \sigma({\bf W}^K \tanh({\bf W}^{K-1} \cdots \tanh({\bf W}^{1} {\bf x}_t))) , $$
where ${\bf x}_t$ are the input feature vectors and $\sigma$ and $\tanh$ denote element-wise logistic and hyperbolic tangent functions. As in the deep NMF experiments, $T=9$ consecutive frames of context are concatenated together, but here the vectors ${\bf x}_t$ are logarithmic magnitude spectra. Thus, the only difference in the input feature representation with respect to deep NMF is the compression of the spectral amplitudes, which is generally considered useful in speech processing, but breaks the linearity assumption of NMF.
 
Previous attempts with DNNs have focused on direct estimation of the clean speech without taking into account the mixture in the output layer, or on direct estimation of a masking function without considering its effect upon the speech estimate. Here, based on our experience with model-based approaches, we train the masking function such that, when applied to the mixture, it best reconstructs the clean speech, which was also proposed in \cite{huang2014ICASSP}. This amounts to optimizing the following objective function for the DNN training:
$$E = \sum_{f,t} ( y_{f,t} m_{f,t} - s^l_{f,t} )^2 = \sum_{f,t} ( \tilde{s}^l_{f,t} - s^l_{f,t} ),$$
where $m$ are the mixture magnitudes and $s^l$ are the speech magnitudes.
Thus, the sequence of output layer activations ${\bf y}_t$ can be interpreted as a time-frequency mask in the magnitude spectral domain, similar to the `Wiener filter' in the output layer of deep NMF \eqref{eq:real_dnmf_wiener}.
In our experiments, this approach leads to 1.5~dB improvements relative to mask estimation. Although this comes from the model-based approach, we include it here so that the DNN results are comparable solely on the context of the deep architecture and not the output layer.

\begin{table}

\caption{DNN source separation performance on the CHiME development set for various topologies.}

\label{tab:dnn_results}
\centering
{\small
\begin{tabular}{c|cccccc|c|c}

SDR [dB] & \multicolumn{6}{c|}{Input SNR [dB]} & & \\

Topology & -6 & -3 & 0 & 3 & 6 & 9 & Avg & \# params \\

3x256 & 3.71 & 5.78 & 7.71 & 9.08 & 10.80 & 12.75 & 8.31 & 644\,K \\

1x1024 & 5.10 & 7.12 & 8.84 & 10.13 & 11.80 & 13.58 & 9.43 & 2.0\,M \\

2x1024 & 5.14 & 7.18 & 8.87 & 10.20 & 11.85 & 13.66 & 9.48 & 3.1\,M \\

3x1024 & 4.75 & 6.74&8.47&9.81 & 11.53&13.38 & 9.11 & 4.1\,M \\

2x1536 & 5.42 & 7.26 & 8.95 & 10.21 & 11.88 & 13.67 & 9.57 & 5.5\,M \\

\end{tabular}
}

\end{table}

Our implementation is based on the open-source software CURRENNT\footnote{http://currennt.sf.net/}. During training, the above objective function is minimized on the CHiME training set, using back-propagation, stochastic gradient descent with momentum, and discriminative layer-wise pre-training. Early stopping based on cross-validation with the CHiME development set, and Gaussian input noise (standard deviation 0.1) are used to prevent aggressive over-optimization on the training set. Unfortunately, our current experiments for deep NMF do not use cross-validation, but despite the advantage this gives to the DNN, as shown below deep NMF nevertheless performs better.
We investigate different DNN topologies (number of layers and number of hidden units per layer) in terms of SDR performance on the CHiME development set. Results are shown in \tablename~\ref{tab:dnn_results}.

\subsection{Baseline 2: sparse NMF}
\label{sec:snmf}
Sparse NMF (SNMF) \cite{Eggert04-SCA} is used as a baseline, by optimizing the training objective, \begin{equation}
\overline{\bf W}^l, \overline{\bf H}^l = \argmin\limits_{{\bf W}^l, {\bf H}^l}  D_{\beta_{2}}({\bf S}^l \mid \widetilde{\bf W}^l {\bf H}^l ) + \mu |{\bf H}^l|_1,
\label{eq:snmf_obj}
\vspace{-0.2cm}
\end{equation}
for each source, $l$.
NMF and which we shall denote by SNMF.
A multiplicative update algorithm to optimize \eqref{eq:snmf_obj} for arbitrary $\beta \geq 0$ is given by \cite{OGrady07-DCS}. During training, we set ${\bf S}^1$ and ${\bf S}^2$ in \eqref{eq:snmf_obj} to the spectrograms of the concatenated noise-free CHiME training set and the corresponding background noise in the multi-condition training set. This yields SNMF bases $\overline{\bf W}^l$, $l=1,2$.
As initial solution for $\overline{\bf W}$, we use exemplar bases sampled at random from the training data for each source.
For the sparsity weight we use $\mu=5$, which performed well for SNMF and DNMF algorithms for both $R^l=100$ and $R^l=1000$ in the experiments of \cite{DNMF}.
In the SNMF experiments, the same basis matrix $\overline{\bf W}$ is used both for determining $\hat{\bf H}$ according to \eqref{eq:sup_nmf_obj} and for reconstruction using \eqref{eq:wiener}.

\subsection{Deep NMF}

In the deep NMF experiments, the KL divergence ($\beta_{1}=1$) is used for the update equations (i.e., in layers $k=1,\dots,K-1$), but we use squared error ($\beta_{2} = 2$) in the discriminative objective \eqref{eq:real_dnmf_wiener} (i.e., in the top layer $k=K$) since this corresponds closely to the SDR evaluation metric, and this combination performed well in \cite{DNMF}.
In all the deep NMF models we initialize the basis sets for all layers using the SNMF bases, $\overline{\bf W}$, trained as described in Section~\ref{sec:snmf}.   We then consider the $C$ last layers to be \emph{discriminatively trained}, for various values of $C$.   This means that we untie the bases for the final $C$ layers (counting the reconstruction layer and analysis layers), and we train the bases $\bW^k$ for $k$ such that $K-C+1\leq k \leq K$
using the multiplicative back-propagation updates described in Section~\ref{sec:app:nmf}.  Thus $C=0$ corresponds to SNMF,  $C\geq1$ corresponds to deep NMF, with the special case $C=1$ previously described as DNMF \cite{DNMF}.

In the experiments, the $K-C$ non-discriminatively trained layers use the full bases $\overline{\bf W}$, which contain multiple context frames.  In contrast the $C$ discriminatively trained layers are restricted to a single frame of context.  This is because the network is being trained to reconstruct a single target frame, whereas using the full context in $\bW^{k}$ and $\bM$ would enforce the additivity constraints across reconstructions of the full context in each layer.  Instead, $\bW^{k>K-C}$ is of size $(F \times R)$, and is initialized to the last $F$ rows of $\overline{\bf W}$ and the matrix $\bM'$, consisting of the last $F$ rows of $\bM$, is used in place of $\bM$.
For deep NMF, the fixed basis functions $\overline{\bf W}$ contain $D_{\mathrm F} = T F R$ parameters that are not trained discriminatively, whereas the final $C$ layers together have $P_{\mathrm D} = C F R$ discriminatively trained parameters, for a total of $P = (T + C)FR$.

\begin{table}
\centering
\caption{deep NMF source separation performance on CHiME Challenge (WSJ-0) development set.}
{ \renewcommand{\tabcolsep}{0.09cm}
\small
\begin{tabular}{l|rrrrrrr|rr}
SDR [dB] & \multicolumn{6}{c}{Input SNR [dB]} &  \\
$R^l=100$    & -6\phantom{.00} & -3\phantom{.00} & 0\phantom{.00} & 3\phantom{.00} & 6\phantom{.00} & 9\phantom{.00} & Avg. & $P_D$\phantom{K} & $P$\phantom{K} \\
\hline
$K=4, \phantom{5} C=0$ (SNMF)& 2.03 & 4.66 & 7.08 & 8.76 & 10.67 & 12.74 & 7.66 & - & 360 K \\
$K=4, \phantom{5} C=1$ (DNMF)& 2.91 & 5.43 & 7.57 & 9.12 & 10.97 & 13.02 & 8.17 & 40 K & 400 K \\
$K=4, \phantom{5} C=2$ & 3.19 & 5.68 & 7.78 & 9.28 & 11.09 & 13.07 & 8.35 & 80 K & 440 K \\
$K=4, \phantom{5} C=3$ & 3.22 & 5.69 & 7.79 & 9.28 & 11.09 & 13.05 & 8.35 & 120 K & 480 K \\
$K=4, \phantom{5} C=4$ & 3.32 & 5.76 & 7.84 & 9.31 & 11.11 & 13.05 & 8.40 & 160 K & 520 K \\
$K=25, C=0$ (SNMF)& 4.16 & 6.46 & 8.51 & 9.90 & 11.61 & 13.40 & 9.01 & - & 360 K \\
$K=25, C=1$ (DNMF)& 4.92 & 7.09 & 8.90 & 10.24 & 12.02 & 13.83 & 9.50 & 40 K & 400 K \\
$K=25, C=2$ & 5.16 & 7.28 & 9.05 & 10.36 & 12.12 & 13.89 & 9.64 & 80 K & 440 K \\
$K=25, C=3$ & 5.30 & 7.38 & 9.14 & 10.43 & 12.18 & 13.93 & 9.73 & 120 K & 480 K \\
$K=25, C=4$ & 5.39 & 7.44 & 9.19 & 10.48 & 12.22 & 13.95 & 9.78 & 160 K & 520 K \\
\\
$R^l=1000$   & -6\phantom{.00} & -3\phantom{.00} & 0\phantom{.00} & 3\phantom{.00} & 6\phantom{.00} & 9\phantom{.00} & Avg. & $P_D$\phantom{K} & $P$\phantom{K} \\
\hline
$K=4, \phantom{5} C=0$ (SNMF)& 1.79 & 4.45 & 6.94 & 8.66 & 10.61 & 12.76 & 7.54 & - & 3.6 M \\
$K=4, \phantom{5} C=1$ (DNMF)& 2.94 & 5.45 & 7.60 & 9.15 & 11.00 & 13.06 & 8.20 & 400 K & 4\phantom{.0} M \\
$K=4, \phantom{5} C=2$ & 3.14 & 5.62 & 7.74 & 9.26 & 11.10 & 13.12 & 8.33 & 800 K & 4.4 M \\
$K=4, \phantom{5} C=3$ & 3.36 & 5.80 & 7.89 & 9.37 & 11.19 & 13.18 & 8.47 & 1.2 M & 4.8 M \\
$K=4, \phantom{5} C=4$ & 3.55 & 5.95 & 8.01 & 9.48 & 11.28 & 13.23 & 8.58 & 1.6 M & 5.2 M \\
$K=25, C=0$ (SNMF)& 4.39 & 6.60 & 8.67 & 10.06 & 11.82 & 13.67 & 9.20 & - & 3.6 M \\
$K=25, C=1$ (DNMF)& 5.74 & 7.75 & 9.55 & 10.82 & 12.55 & 14.35 & 10.13 & 400 K & 4\phantom{.0} M \\
$K=25, C=2$ & 5.80 & 7.80 & 9.59 & 10.86 & 12.59 & 14.39 & 10.17 & 800 K & 4.4 M \\
$K=25, C=3$ & 5.84 & 7.82 & 9.62 & 10.89 & 12.61 & 14.40 & 10.20 & 1.2 M & 4.8 M 
\end{tabular}
}
\label{tab:chime_dev}
\end{table}

\section{Discussion}
Results in terms of SDR are shown for the experiments using DNNs in Table~\ref{tab:dnn_results}, and for the deep NMF family in Table~\ref{tab:chime_dev}, for a range of topologies.   The deep NMF framework yields strong improvements relative to SNMF. Comparing the DNN and deep NMF approaches, the best deep NMF topology achieves an SDR of 10.20~dB, outperforming the best DNN result of 9.57~dB, for a comparable number of parameters (4.8M for deep NMF versus 5.5M for the DNN).
The smallest deep NMF topology that outperforms the best DNN topology, is obtained for $R^l=100, K=25, C=2,$ and achieves an SDR of 9.64~dB using at least an order of magnitude fewer parameters (only 440K parameters, only 80K of which are discriminatively trained).  

Analyzing further the effect of topology on performance for deep NMF, discriminative training of the first layer gives the biggest improvement, but training more and more layers consistently improves performance, \longtext{especially in low SNR conditions,} while only adding a modest number of parameters per layer. 
\longtext{Moving from $R^l=100$ to $R^l=1000$  does not lead to as much gain as one might expect despite the huge increase in parameter size. This could be because we are currently only training on 10~\% 
of the data, and used a fairly conservative convergence criterion. }
For the same model size, using $K=25$ layers leads to large gains in performance without increasing training time and complexity. However, it comes at the price of an increased computation cost at inference time. Intermediate topology regimes need to be further explored to get a better sense of the best speed/accuracy trade-off.

As potential further work in the speech enhancement domain, we could consider investigating the application of our framework to models with continuity constraints or factorial structure \cite{OzerovVB12,mysore2012variational,Fevotte2013ICASSP05,Simsekli2014ICASSP05}.
More general future research directions within the deep unfolding paradigm include experiments with unfolding of other inference algorithms, such as loopy belief propagation \longtext{for Markov random fields } or variational inference algorithms for intractable generative models.

In conclusion, a general framework was introduced that allows model-based approaches to guide the exploration of the space of deep network architectures, which would otherwise be difficult to navigate.  
We have shown how conventional sigmoid networks could be seen as unfolded mean-field inference in Markov random fields, leading to possible generalizations to other inference algorithms such as belief propagation and its variants.   Finally, we showed how model-based problem constraints of non-negative matrix factorization can be incorporated via deep unfolding into a novel deep architecture.  
By reasoning at the problem level with the model-based approach, our methodology allows one to derive inference architectures and training methods that otherwise would be difficult to obtain.   We hope that this framework will inspire a new generation of novel deep network architectures suitable for tackling difficult problems that require high-level domain insights.

\newpage

\bibliographystyle{IEEEbib}
\bibliography{refs}

\newpage
\appendixpage
\appendix

\section{Derivations for parameters of binary MRFs}
\label{sec:binaryMRF}
Here we show how the parameters of a sigmoid network can be derived from binary MRFs, and conversely how an equivalent set of MRF parameters can be extracted from the sigmoid parameters.  
\subsection{Binary MRF to sigmoid network parameters}
Consider the binary MRF case with $h_i, v_l \in \{0,1\}$, and $i,j \in  \mathcal{I}_{\mathsf{h}} = \{1,  \cdots, N_{\mathsf{h}} \}$,  and $l \in \mathcal{I}_{\mathsf{v}} = \{1,  \cdots, N_{\mathsf{v}} \}$ using the $2\times 2$ log potential matrices $\psi^{(i,j)}$, and elements,
\begin{align}
\psi^{(i,j)}_{h_i,h_j} = 
\begin{cases}
\Psi(h_i, h_j): & \mbox{if } (i,j) \in \mathcal{E}^{\mathsf{hh}}
\\
0: & \mbox{otherwise},
\end{cases} 
\end{align} and similarly for matrices $\psi^{(i,l)}$,
so that we can write,  
\begin{align}
\log p(h|v) 
&\logpropto 
			\sum_{(i,j) \in \mathcal{E}_{\mathsf{hh}}}
				\psi^{(i,j)}_{h_i,h_j} 
			+ \sum_{(i,l) \in \mathcal{E}_{\mathsf{hv}}} 
				\psi^{(i,l)}_{h_i,v_l}
\nonumber \\
&\logpropto 
			\sum_{(i,j) \in \mathcal{E}_{\mathsf{hh}}}
			\Bigl(
				h_i h_j \psi^{(i,j)}_{1,1} 
			+ (1-h_i) h_j \psi^{(i,j)}_{0,1} 
			+ h_i (1-h_j) \psi^{(i,j)}_{1,0}
			+ (1-h_i) (1-h_j) \psi^{(i,j)}_{0,0} 
			\Bigr)
\nonumber \\
	& \quad + \sum_{(i,l) \in \mathcal{E}_{\mathsf{hv}}} 
			\Bigl(
				h_i v_l \psi^{(i,l)}_{1,1} 
			+ (1-h_i) v_l \psi^{(i,l)}_{0,1} 
			+ h_i (1-v_l) \psi^{(i,l)}_{1,0}
			+ (1-h_i) (1-v_l)v_l \psi^{(i,l)}_{0,0} 
			\Bigr)
\nonumber \\
&\logpropto 
			\sum_{(i,j) \in \mathcal{E}_{\mathsf{hh}}}
			\Biggl[
			h_i h_j 
				\Bigl(\psi^{(i,j)}_{1,1} 
					- \psi^{(i,j)}_{0,1} 
					- \psi^{(i,j)}_{1,0} 
					+ \psi^{(i,j)}_{0,0}
				\Bigr)
			+ h_i 
				\Bigl(\psi^{(i,j)}_{1,0} 
					- \psi^{(i,j)}_{0,0}
				\Bigr)
			+ h_j 
				\Bigl(\psi^{(i,j)}_{0,1} 
					- \psi^{(i,j)}_{0,0}
				\Bigr)
			+ \psi^{(i,j)}_{0,0}
			\Biggr]			
\nonumber \\
	& \quad + \sum_{(i,l) \in \mathcal{E}_{\mathsf{hv}}} 
			\Biggl[
			h_i v_l 
				\Bigl(\psi^{(i,l)}_{1,1} 
					- \psi^{(i,l)}_{0,1} 
					- \psi^{(i,l)}_{1,0} 
					+ \psi^{(i,l)}_{0,0}
				\Bigr)
			+ h_i 
				\Bigl(\psi^{(i,l)}_{1,0} 
					- \psi^{(i,l)}_{0,0}
				\Bigr)
			+ v_l 
				\Bigl(\psi^{(i,l)}_{0,1} 
					- \psi^{(i,l)}_{0,0}
				\Bigr)
			+ \psi^{(i,l)}_{0,0}
			\Biggr]			
\nonumber \\
&\logpropto 
			{\textstyle \frac{1}{2}}h^T A h +  h^T b + h^T C v,
\nonumber \\
\end{align}
where $a \logpropto b$ implies $a = b + c$, for some constant $c$ that does not depend on $h$,  and where
\begin{align}
\nonumber \\
h & \defeq 	[h_1, \cdots, h_{N_{\mathsf{h}}}]^T
,\quad
& v  &\defeq 	[v_1, \cdots, v_{N_{\mathsf{v}}}]^T
\nonumber \\
a_{i,j} & \defeq 
			\Bigl(		
				\psi^{(i,j)}_{1,1} 
				- \psi^{(i,j)}_{0,1} 
				- \psi^{(i,j)}_{1,0} 
				+ \psi^{(i,j)}_{0,0}
			\Bigr)
,
& c_{i,l} &\defeq
			\Bigl(		
				\psi^{(i,l)}_{1,1} 
				- \psi^{(i,l)}_{0,1} 
				- \psi^{(i,l)}_{1,0} 
				+ \psi^{(i,l)}_{0,0}
			\Bigr)
,
\nonumber \\
b_{i}& \defeq   \sum_{j\in \mathcal{N}^{\mathsf{hh}}_i}				
				\Bigl(\psi^{(i,j)}_{1,0} 
					- \psi^{(i,j)}_{0,0}
				\Bigr)
				+ \sum_{l\in \mathcal{N}^{\mathsf{hv}}_i}				
				\Bigl(\psi^{(i,l)}_{1,0} 
					- \psi^{(i,l)}_{0,0}
				\Bigr)
.
\label{eq:mrf_to_sigmoid}
\end{align}
The factor of $1/2$ is necessary in $h^T A h$ since $a_{i,j} = a_{j,i}$, and each edge potential is counted twice in $h^T A h$, but only once in the sum over edges ${(i,j) \in \mathcal{E}_{\mathsf{hh}}}$.   Note that all terms that do not depend on $h$ are dropped, since they do not affect the posterior $p(h|v)$.  

\subsection{Sigmoid network to binary MRF parameters}
It is easy to go in the other direction to derive MRF parameters $\psi$ from sigmoid network parameters $A$, $b$, and $C$.  Using the same model defined above,
\begin{align}
\log p(h|v) &\logpropto
	{\textstyle \frac{1}{2}} h^T A h  + h^T b  + h^T C v 
\nonumber \\
&\logpropto
	\sum_{i,j \in \mathcal{I}_{\mathsf{h}}} 
		{\textstyle \frac{1}{2}}  h_i  h_j a_{i,j}
	+  \sum_{i \in \mathcal{I}_{\mathsf{h}}} 
		h_i b_i 
	+ \sum_{i \in \mathcal{I}_{\mathsf{h}} ,l \in \mathcal{I}_{\mathsf{v}}} 
		h_i  v_l c_{i,l}
\nonumber \\
&\logpropto
			\sum_{(i,j) \in \mathcal{E}_{\mathsf{hh}}}
			\Bigl(
			h_i h_j a_{i,j}
			+ h_i \frac{b_i}{n_i}  
			+ h_j \frac{b_j}{n_j}  
			\Bigr)			
			+ \sum_{(i,l) \in \mathcal{E}_{\mathsf{hv}}}
			h_i v_l c_{i,l},
\nonumber \\
&\logpropto
			\sum_{(i,j) \in \mathcal{E}_{\mathsf{hh}}}
			\psi^{(i,j)}_{h_i,h_j} 
			+\sum_{(i,l) \in \mathcal{E}_{\mathsf{hv}}}
		    \psi^{(i,l)}_{h_i,v_l},			
\end{align}
where the values of $\psi$ for $(i,j) \in \mathcal{E}_{\mathsf{hh}}$ and for  $(i,l) \in \mathcal{E}_{\mathsf{hv}}$  are given by
\begin{align}
\psi^{(i,j)}_{0,0} & = 0
,
& \psi^{(i,j)}_{0,1} & = \frac{b_j }{n_j} 
,
& \psi^{(i,l)}_{0,0} & = 0
,
& \psi^{(i,l)}_{0,1} & = 0
,
\nonumber \\
\psi^{(i,j)}_{1,0} & = \frac{b_i }{n_i} 
,
& \psi^{(i,j)}_{1,1} & = a_{i,j} + \frac{b_i }{n_i} + \frac{b_j }{n_j} 
,
& \psi^{(i,l)}_{1,0} & = 0
,
& \psi^{(i,l)}_{1,1}& = c_{i,j}.
\end{align}
We drop the factor of $1/2$ and normalize by the degree $n_i = |\mathcal{N}^{\mathsf{hh}}_i|$ of node $i$ to go from summing over all $i$ to summing over edges $(i,j)$.  
Note that any constants of proportionality that we may add to $\psi^{(i,l)}$ and $\psi^{(i,l)}$ will factor out in the end.  Also, here we assign the bias terms $b_i$ arbitrarily to $\psi^{(i,j)}$, although in \eqref{eq:mrf_to_sigmoid} we added together the bias contributions from both $\psi^{(i,j)}$ and $\psi^{(i,l)}$ to form $b_i$ rather than keeping them separate.  
This has no effect on $p(h|v)$ since those terms are independent of $v_l$.   

Finally there is the question of what our generalized activation function looks like as a function of $a_{i,j}$, $b_i$, and $c_{i,l}$.
Here we leave out the edge $i,j$ indexing on the exponent $\lambda$, and iteration/layer $k$.  We also define $\mu_i = q(h_i=1)$ and 	$\msga{\mu}{i}{j} =  \msg{i}{j}(h_j=1)$.
We start with the message, 
\begin{align}
\msga{m}{j}{i}(h_i) 
&\propto 
	\Biggl(
		\sum_{h_j} 
		\frac{q(h_j)}
		{ \msgx{i}{j}{\lambda}(h_j)}
		e^{\lambda \Psi(h_i,h_j)}
	\Biggr)^{\frac{1}{\lambda}}	
\nonumber \\	
&\propto 
	\Biggl(
		\frac{\mu_j}{\msgax{\mu}{i}{j}{\lambda}}
		e^{\lambda \psi^{(i,j)}_{h_i,1} }
		+ \frac{1 - \mu_j}
		{(1 - \msga{\mu}{i}{j})^{\lambda}}
		e^{ \lambda h_i \psi^{(i,j)}_{h_i,0} }
	\Biggr)^{\frac{1}{\lambda}}.
\label{eq:generalized-mf-bp-binary}
\end{align}
Combining this with the belief update yields
\begin{align}
q(h_i)
&\propto  
\exp\Biggl( 
		\sum_{j \in \mathcal{N}^{\mathsf{hh}}_i}		
		\log \msg{j}{i} (h_i)
		+ \sum_{l \in \mathcal{N}^{\mathsf{hv}}_i}
		\psi^{(i,l)}_{h_i,v_l^{\ast}}
	\Biggr)
\nonumber \\	
\mu_i
&=   
\operatorname{logistic}
\Bigl( 
	\sum_{j \in \mathcal{N}^{\mathsf{hh}}_i}		
	\bigl(
	\log \msga{m}{j}{i}(1)
	-\log \msga{m}{j}{i}(0)					
	\bigr)
	+ \sum_{l \in \mathcal{N}^{\mathsf{hv}}_i}
	\bigl(
		\psi^{(i,l)}_{1,v_l^{\ast}} 
		- 	
		\psi^{(i,l)}_{0,v_l^{\ast}} 	
	\bigr)				
\Bigr)
\nonumber \\	
&=
\operatorname{logistic}
	\left( 
		{\frac{1}{\lambda}}	
		\sum_{j \in \mathcal{N}^{\mathsf{hh}}_i}		
		\log 
		\frac{
			\frac{\mu_j}{ \msgax{\mu}{i}{j}{\lambda}}
			e^{\lambda \psi^{(i,j)}_{1,1} }
			+ 
			\frac{1 - \mu_j}{(1 - \msga{\mu}{i}{j})^{\lambda}}
			e^{ \lambda \psi^{(i,j)}_{1,0} }
		}
		{
			\frac{\mu_j}{ \msgax{\mu}{i}{j}{\lambda}}
			e^{\lambda \psi^{(i,j)}_{0,1} }
			+ 
			\frac{1 - \mu_j}{(1 - \msga{\mu}{i}{j})^{\lambda}}
			e^{ \lambda \psi^{(i,j)}_{0,0} }
		}
		+ \sum_{l \in \mathcal{N}^{\mathsf{hv}}_i}
			\psi^{(i,l)}_{1,v_l^{\ast}} 
			- 	
			\psi^{(i,l)}_{0,v_l^{\ast}} 	
	\right)
\nonumber \\
&= 
\operatorname{logistic}
	\left( 
		{\frac{1}{\lambda}}	
		\sum_{j \in \mathcal{N}^{\mathsf{hh}}_i}		
		\log 
		\frac{
			\frac{\mu_j}{ \msgax{\mu}{i}{j}{\lambda}}
			e^{\lambda ( a_{i,j}   +  \frac{b_i }{n_i} +  \frac{b_j }{n_j} ) }
			+ 
			\frac{1 - \mu_j}{(1 - \msga{\mu}{i}{j})^{\lambda}}
			e^{\lambda ( \frac{b_i }{n_i} ) }
		}
		{
			\frac{\mu_j}{ \msgax{\mu}{i}{j}{\lambda}}
			e^{\lambda ( \frac{b_j }{n_j} ) }
			+ 
			\frac{1 - \mu_j}{(1 - \msga{\mu}{i}{j})^{\lambda}}						
		}
		+ \sum_{l \in \mathcal{N}^{\mathsf{hv}}_i}
		c_{i,l} v_l^{\ast} 	
	\right)
.
\label{eq:bpbelief_binary}
\end{align}
In a feedforward network with untied parameters, the backward-going message $ \msg{i}{j}{\lambda}(h_j)$ can be considered uniform ($\msga{\mu}{i}{j} = 1/2$) in which case it cancels, leaving 
\begin{align}
\mu_i
&=
\operatorname{logistic}
	\left( 
		{\frac{1}{\lambda}}	
		\sum_{j \in \mathcal{N}^{\mathsf{hh}}_i}		
		\log 
		\frac{
			\mu_j 
			e^{\lambda ( a_{i,j}   +  \frac{b_i }{n_i} +  \frac{b_j }{n_j} ) }
			+ 
			(1 - \mu_j)
				e^{ \lambda (\frac{b_i }{n_i} ) }
		}
		{
			\mu_j 
			e^{\lambda (\frac{b_j }{n_j} ) }
			+ 
			(1 - \mu_j) 
		}
		+ \sum_{l \in \mathcal{N}^{\mathsf{hv}}_i}
		c_{i,l} v_l^{\ast} 	
	\right)
.
\label{eq:bpbelief_binary_abc}
\end{align}

It is easy to verify that in the limit as $\lambda \rightarrow 0$, both \eqref{eq:bpbelief_binary} and \eqref{eq:bpbelief_binary_abc} become the standard sigmoid activation function.   Applying the power-mean limit to the log of the numerator in \eqref{eq:bpbelief_binary} we get 
\begin{align}
&\lim_{\lambda\rightarrow 0}
	{\frac{1}{\lambda}}
	\log
		\left(
			\mu_j 
			e^{\lambda( 
				a_{i,j}   +  \frac{b_i }{n_i} +  \frac{b_j }{n_j} 
				- \log {\msga{\mu}{i}{j}} )
				}
			+ 
			(1 - \mu_j)
				e^{\lambda ( \frac{b_i }{n_i} - \log ({1 - \msga{\mu}{i}{j}})) }					
		\right)			
\nonumber \\
&=	\mu_j 
		( 
			a_{i,j}   +  \frac{b_i }{n_i} +  \frac{b_j }{n_j} 
			- \log {\msga{\mu}{i}{j}} 
		)
		+ 
		(1 - \mu_j)
		(
		\frac{b_i }{n_i} - \log ({1 - \msga{\mu}{i}{j}})
		)					
,
\end{align}
and for the log of the denominator we have
\begin{align}
&\lim_{\lambda\rightarrow 0}
	{\frac{1}{\lambda}}	
	\log 
	( 
		\mu_j  
		e^{\lambda ( \frac{b_j }{n_j}  - \log \msga{\mu}{i}{j} ) }
		+
		(1 - \mu_j)
		e^{\lambda \bigl( - \log (1 - \msga{\mu}{i}{j})\bigr)}				
	)
\nonumber \\
&=	\mu_j ( \frac{b_j }{n_j}   - \log \msga{\mu}{i}{j} )
-  (1 - \mu_j)   \log (1 - \msga{\mu}{i}{j})		
.
\end{align}
Substituting their difference back into \eqref{eq:bpbelief_binary} and canceling terms we get
\begin{align}
\mu_i 
&= 
\operatorname{logistic}
	\Biggl(
		\sum_{j \in \mathcal{N}^{\mathsf{hh}}_i}		
		\left(
			a_{i,j}  \mu_j 
			+ 
			\frac{b_i }{n_i} 
		\right)
		+ \sum_{l \in \mathcal{N}^{\mathsf{hv}}_i}
		c_{i,l} v_l^{\ast} 	
	\Biggr)
\nonumber \\
&= 
\operatorname{logistic}
	\left(
		A \mu 
		+ 
		b
		+C v^{\ast} 	
	\right),
\label{eq:bpbelief_binary_limit}
\end{align}
where we use the fact that $\sum_{j \in \mathcal{N}^{\mathsf{hh}}_i} 1 = n_i$.

\section{Review of power mean and log domain power mean}
\label{sec:mean}
The generalized power mean, with values $x_i > 0 $ with weights $w_i \geq 0$ such that $\sum_i w_i = 1$, and exponent $a$, is
\begin{align}
M_{a}(w,x) &= \Big(\sum_i w_i x_i^a \Big)^{1/a}.
\end{align}
It is also useful to know its equivalent log domain form, with $z = \log(x)$, $L = \log M$
\begin{align}
L_{a}(w,z) &= \frac{1}{a}\log\Bigl(\sum_i w_i e^{a z_i} \Bigr).
\end{align}

Depending on the exponent, $a$, it gives rise to various commonly used means:

\bgroup
\def\arraystretch{1.5}
\begin{tabular}{|c|c|c|c|}
\hline general mean & a &  $M_{a}(x,w)$ &  $L_{a}(z,w) $\\ 
\hline min & $a\rightarrow -\infty$ & $\min_i { x_i}$ & $\min_i  z_i$  \\ 
\hline harmonic mean & $a=-1$ &  $1 \big/ \sum_i \frac{w_i}{ x_i} $ & $ - \log \sum_i {w_i}e^{-z_i}$ \\ 
\hline geometric mean  & $a \rightarrow 0$ & $\prod_i x_i^{w_i}$ & $\sum_i w_i z_i$ \\ 
\hline arithmetic mean  & $a=1$ &  $\sum_i w_i x_i$ & $\log\Bigl(\sum_i w_i e^{z_i} \Bigr)$ \\ 
\hline max  & $a\rightarrow \infty$ & $\max_i { x_i}$ & $\max_i z_i $\\ 
\hline 
\end{tabular} 
\egroup

The limit of the improper form with $a\rightarrow 0$ for the geometric mean can be derived as a form of the fundamental theorem of calculus as follows:
\begin{align}
\lim_{a \rightarrow 0} \log L_{a}(w,z) 
&= \lim_{a \rightarrow 0}  \frac{1}{a}\left[ \log\Bigl(\sum_i w_i e^{a z_i} \Bigr) - \log\Bigl(\sum_i w_i e^{(0) z_i} \Bigr)\right]
\nonumber\\
\nonumber\\
&= \Bigl.\frac{\partial}{\partial a} \log\Bigl(\sum_i w_i e^{a z_i}\Bigr) \Bigr|_{a=0}
= \Bigl.\frac{\sum_i w_i z_i e^{a z_i}}
       	{\sum_i w_i e^{a z_i}} \Bigr|_{a=0}        	
= \sum_i w_i z_i; 
\nonumber\\
\lim_{a \rightarrow 0} M_{a}(w,x) &= \exp\Big(\sum_i w_i \log(x_i)\Big) = \prod_i x_i^{w_i}.
\end{align}

The power mean inequality holds that $a\leq b$ implies
$
M_{a}(w,x) \leq M_{b}(w,x),
$
which is proved using Jensen's inequality.  Assuming that $0< a\leq b$,
\begin{align}
\Bigl(M_{a}(w,x)\Bigr)^{b} =  \Big(\sum_i w_i x_i^a \Big)^{b/a} &\leq  \sum_i w_i x_i^b  \quad \text{by Jensen's inequality and convexity of $(\cdot)^{b/a}$}
\nonumber
\\
\Big(\sum_i w_i x_i^a \Big)^{1/a}  & \leq  \Big(\sum_i w_i x_i^b \Big)^{1/b} \; \text{applying increasing function $(\cdot)^{1/b}$ to both sides.}  
\end{align}
In the case of $a<b<0$ we can proceed as above with $\bigl(M_{a}(w,x)\bigr)^{-b}$, so that $(\cdot)^{-b/a}$ is convex, and $(\cdot)^{-1/b}$ is increasing, and we produce the same inequality.  For $a=0$ or $b=0$, the continuity of the limiting case completes the proof.

\section{Generalized message forms for mean field and belief propagation}
\label{sec:generalized_forms}
We consider three forms for message passing, here for clarity we leave out the edge $i,j$ and iteration $k$ indexing on the exponent $\lambda$ and $\kappa$. 
The first interpolates between BP ($\lambda=1$) and MF ($\lambda\rightarrow 0$):
\begin{align}
\msgax{\tilde{m}}{j}{i}{k}(h_i) \propto 
	\Bigl(
		\sum_{h_j} 
		q^{k-1}(h_j) \Big(\frac{e^{\Psi^{k}(h_i,h_j)}}{\msgax{\tilde{m}}{j}{i}{k-1}(h_j)}\Big)^{\lambda}		
	\Bigr)^{\frac{1}{\lambda}}
.		
\label{eq:generalized-mf-bp}
\end{align}

The second form interpolates between sum-product BP $\kappa = 1$ and max-product BP $\kappa \rightarrow \infty$:
\begin{align}
\msgax{\tilde{m}}{j}{i}{k}(h_i) \propto 
	\Bigl(
		\sum_{h_j} 
		\frac{1}{N_{\mathsf{h_j}}}\Big(q^{k-1}(h_j) \frac{e^{\Psi^{k}(h_i,h_j)}}{\msgax{\tilde{m}}{j}{i}{k-1}(h_j)}\Big)^{\kappa}		
	\Bigr)^{1/\kappa}
.
\label{eq:generalized-sumprod-maxprod}
\end{align}

A third form combines both to produce sum-product BP ($\lambda=1$, $\kappa = 1$), max-product BP ($\lambda=1$, $\kappa\rightarrow \infty$), and mean field MF ($\lambda\rightarrow0$, $\kappa = 1$ ):
\begin{align}
\msgax{\tilde{m}}{j}{i}{k}(h_i) \propto 
	\Biggl(
		\sum_{h_j} 
      \frac{q^{k-1}(h_j)^{\kappa}}{N_{\mathsf{h_j}}^{\lambda}} \Bigl(\frac{e^{\lambda\kappa\Psi^{k}(h_i,h_j)}}{\msgax{\tilde{m}}{j}{i}{k-1}(h_j)}\Bigr)^{\lambda\kappa}
	\Biggr)^{\frac{1}{\lambda\kappa}}	
.	
\label{eq:generalized-combo}
\end{align}

\section{Derivations for deep NMF}
\subsection{Chain rule for Deep NMF}
In the case of DNMF, the architecture is defined as follows, with $\bH^{0}$ initialized, e.g., randomly:
\begin{align}
\bH^{k+1} =&   f(\bW^k , \bM , \bH^k) \\
\hat{\bS} =& g(\bW^K, \bM , \bH^K).
\end{align}
The objective function is given by:
\begin{equation}
\mathcal{E} = \sum_l \gamma_l D_\beta(\bS^l |  \widehat \bS^l).
\end{equation}

We assume that the sets of basis functions for all sources are stacked horizontally, and their activations are stacked vertically. In order to be able to assume different numbers of basis functions $R_l$ for each source, the stacked structures are indexed using a single index $r$, such that the elements corresponding to source $l$ are those for which $r \in I_l$, where we defined the set $I_l = [\![ i_l, i_{l+1}[\![$ of integers $r$ such that $i_l \leq r < i_{l+1}$, with $i_1 = 1$ and $i_{l+1} = i_{l} + R_l$.

We compute the gradient of $\mathcal{E}$ with respect to the parameters in the $k$-th layer, $ w^k_{n,l,r}$.

For $k=K$, the gradient $\frac{\partial \mathcal{E}}{\partial w^K_{n,r}}$ is the one obtained for the reconstruction basis functions in Discriminative NMF.

For $k<K$, we use the chain rule to get:
\begin{align}
&\hspace{-2cm}  \frac{\partial \mathcal{E}}{\partial w^k_{n,r}} = \sum_{l'} \gamma_{l'} \frac{\partial D_\beta(\bS^{l'}|\hat \bS^{l'})}{\partial w^k_{n,r}} \notag \\
&\hspace{-2cm} =
\sum_t
\sum_{r_{k+1}}\Bigg ( \dots \Bigg (
\sum_{r_{K}} \Bigg (
\sum_{n'} \left(
\sum_{l'} \gamma_{l'} \frac{\partial D_\beta(S^{l'}_{n',t}|\hat S^{l'}_{n',t})}{\partial \hat  S^{l'}_{n',t}}\right )
\frac{\partial g^l_{n',t}(\bW^K,\bM,\bH^K)}{\partial h^K_{r_K,t}}  \Bigg )
\frac{\partial f_{r_K,t}(\bW^{K-1},\bM,\bH^{K-1})}{\partial h^{K-1}_{r_{K-1},t}} \Bigg ) \dots \notag \\
& \hspace{7cm}\dots   \frac{\partial f_{r_{k+2},t }(\bW^{k+1},\bM,\bH^{k+1}) }{\partial h^{k+1}_{r_{k+1},t } }   \Bigg)
\frac{\partial f_{r_{k+1},t }(\bW^k,\bM,\bH^k) }{\partial w^k_{n,r}}.  \notag
\end{align}
This can be rewritten slightly more clearly as:
\begin{align}
\frac{\partial \mathcal{E}}{\partial w^k_{n,r}} & =
\sum_t
\sum_{r_{k+1}} \frac{\partial f_{r_{k+1},t }(\bW^k,\bM,\bH^k) }{\partial w^k_{n,r}}
\Bigg ( \sum_{r_{k+2}} \frac{\partial f_{r_{k+2},t }(\bW^{k+1},\bM,\bH^{k+1}) }{\partial h^{k+1}_{r_{k+1},t } }
\Bigg (  \dots \notag \\
& \dots \sum_{r_{K}} \frac{\partial f_{r_K,t}(\bW^{K-1},\bM,\bH^{K-1})}{\partial h^{K-1}_{r_{K-1},t}} \Bigg (
\sum_{n'} \left(
\sum_{l'} \gamma_{l'} \frac{\partial D_\beta(S^{l'}_{n',t}|\hat S^{l'}_{n',t})}{\partial \hat  S^{l'}_{n',t}}\right )
\frac{\partial g^l_{n',t}(\bW^K,\bM,\bH^K)}{\partial h^K_{r_K,t}}  \Bigg )
 \Bigg ) \dots \Bigg). \notag
\end{align}

Once the values of $h^k$ have been computed in a forward pass, the gradient can be computed in a backward pass, starting from the top layer, which is the most inner term in the sum above.

We give the expression for all terms in the sum, starting from the inside.

\subsection{Top ($K$-th) layer derivative w.r.t.\ $\bH^K$ with Least-Squares divergence measure}

Even though we typically use KL-divergence ($\beta=1$) to train the analysis basis functions and estimate the activations $\bH$, nothing prevents us from using a different discrepancy measure at the final layer. If we want to optimize the signal-to-noise ratio (SNR) in the case where features are magnitude spectra, we can minimize the Euclidean distance $D_2^l := D_2({\bf S}^l | \hat{\bf S}^l)$ between the (true) magnitude spectrum of the source $l$, ${\bf S}^l$, and that of its reconstruction, $\hat{\bf S}^l$. This indeed directly corresponds to maximizing the SNR, neglecting the difference between noisy and oracle phases (we thus optimize for an upper-bound of the actual SNR). Note that similar updates can be obtained for $\beta=1$ in the last layer as well.
We are also only interested in the quality of the reconstruction of a single source, typically speech in a speech enhancement scenario.

Let us thus assume that $\gamma_l =1, \gamma_{l',l'\neq l} =0$, $\beta =2 $ in the reconstruction layer and Wiener filter is used for reconstruction (and that fact is included in the optimization), i.e., we optimize to reconstruct a single source $l$ (e.g., speech), using the L2 norm as the error measure, and assume that the source is reconstructed using
$$g^l(\bW^K, \bM , \bH^K) =  \frac{\bW^{K,l} \bH^{K,l}}{\bW^K \bH^K} \circ \bM,$$
where ${\bH^{K,l}} = [{\bf h}^{K}_{i_l}; \cdots; {\bf h}^{K}_{i_{l+1}-1}]$ (we use the notation $[ {\bf a}; {\bf b} ]$ for $[ {\bf a}^\intercal {\bf b}^\intercal ]^\intercal$) and ${\bW^{K,l}}$ is defined accordingly. Let us denote ${\bf \Lambda} = \bW^K \bH^K = \sum_l \bW^{K,l} \bH^{K,l}$, ${\bf \Lambda}^l = \bW^{K,l} \bH^{K,l}$, and ${\bf \Lambda}^{\overline{l}} = {\bf \Lambda} - {\bf \Lambda}^l$.
Then we have:
\begin{align}
\left[\nabla_{\bH^{K,l}}{\mathcal{E}}\right]_+ &=  {{\bW}^{K,l}}^\intercal
         \frac{ {\bf M}^2  \circ {\bf \Lambda}^l \circ {\bf \Lambda}^{\overline{l}} } { {\bf \Lambda}^3 },  \\
\left[\nabla_{\bH^{K,l}}{\mathcal{E}}\right]_- &= {{\bW}^{K,l}}^\intercal
         \frac{ {\bf M} \circ {\bf S}^l  \circ {\bf \Lambda}^{\overline{l}} } { {\bf \Lambda}^2 },   \\
\left[\nabla_{\bH^{K,\overline{l}}}{\mathcal{E}}\right]_+ &= {{\bW}^{K,\overline{l}}}^\intercal
         \frac{ {\bf M} \circ {\bf S}^l \circ {\bf \Lambda}^l } { {\bf \Lambda}^2 } , \\
\left[\nabla_{\bH^{K,\overline{l}}}{\mathcal{E}}\right]_- &= {{\bW}^{K,\overline{l}}}^\intercal
         \frac{ {\bM}^2  \circ ({\bf \Lambda}^l)^2 } { {\bf \Lambda}^3 },
\end{align}
where $\bH^{K,\overline{l}} = [{\bH}^{K,1}; \cdots; {\bH}^{K,l-1}; {\bH}^{K,l+1}; \cdots; {\bH}^{S} ] $, and $\bW^{K,\overline{l}}$ is defined accordingly.

\subsection{Intermediate ($k+1$-th) layer derivative w.r.t.\ $\bH^k$ and $\bW^k$, $k<K$}

Even though we may be using any $\beta$ divergence for the last layer, and we in fact use $\beta=2$, i.e., the L2 distance, we use $\beta=1$, i.e., the KL divergence, for the intermediate layers. We noticed that this combination leads to better results than using the same $\beta$ all the way through. This may be due to the KL divergence being better suited to decomposing mixtures, as was shown in previous evaluations.

To simplify the update equations for $\bW^k$, we assume that $\bW^k$ is not normalized in the final layers where it is optimized. However, we do assume that it is normalized in lower layers which are not optimized: the analysis basis functions were indeed trained under that assumption to avoid obtaining trivial solutions regarding sparsity simply by rescaling $\bW$ and $\bH$. Actually, any layer where $\bW^k$ is not optimized can use the normalized versions of the updates without any change.

For the layers where $\bW^k$ is not normalized, the update equation for $\bH^{k+1}$, which determines the structure of the intermediate layers, is given by:
\begin{equation}
h^{k+1}_{r_{k+1},t}(\bW^k , \bM , \bH^k) = h^k_{r_{k+1},t} \frac{\displaystyle \sum_n {w}^k_{n,r_{k+1}} \frac{m_{n,t}}{\sum_{r'} {w}^k_{n,r'} h^k_{r',t}} }
{\displaystyle \sum_n {w}^k_{n,r_{k+1}} + \mu}.
\end{equation}
Let $\alpha_r^k = \sum_n {w}^k_{n,r}$ and $\Lambda^k_{n,t} = \sum_{r'}   w^k_{n,r'} h^k_{r',t}$.
The derivative of the intermediate layers is obtained as:
\begin{align}
\frac{\partial h^{k+1}_{r_{k+1},t}}{\partial h^k_{r_k,t}} =& \frac{ 1 }{\displaystyle \alpha^k_{r_k} + \mu}
 \sum_n {w}^k_{n,r_k} \frac{m_{n,t}}{\Lambda^k_{n,t}}
- \frac{ h^k_{r_k,t} }{\displaystyle \alpha^k_{r_k} +\mu}  \sum_n   (w_{n,r_k}^k)^2 \frac{ m_{n,t}}{ (\Lambda_{n,t}^k)^2 }    ,\quad \text{for}\; r_{k+1} = r_k\\
\frac{\partial h^{k+1}_{r_{k+1},t}}{\partial h^k_{r_k,t}} =& - \frac{ h^k_{r_{k+1},t} } {\displaystyle \alpha^k_{r_{k+1}}+\mu}   \displaystyle \sum_n  w^k_{n,r_{k+1}}  w^k_{n,r_k}  \frac{  m_{n,t}}{ (\Lambda_{n,t}^k)^2 }   ,\quad \text{for}\; r_{k+1} \neq r_k
\end{align}
and
\begin{align}
\frac{\partial h^{k+1}_{r_{k+1},t}}{\partial w^k_{n_0, r_k}} &= \frac{h^k_{r_k,t}}{\alpha^k_{r_k} +\mu} \left( 1 - \frac{w^k_{n_0,r_k}    h^k_{r_k,t}}{\Lambda^k_{n_0,t}} \right) \frac{m_{n_0,t}}{ \Lambda^k_{n_0,t} }
- \frac{ h^k_{r_k,t}}{\displaystyle( \alpha^k_{r_k} +\mu )^2} \left( \sum_n  w^k_{n,r_k} \frac{m_{n,t}}{ \Lambda^k_{n,t} }  \right)  ,\quad \text{for}\; r_{k+1} = r_k \notag \\
\frac{\partial h^{k+1}_{k+1,t}}{\partial w^k_{n_0, r_k}} &= - \frac{h^k_{r_k,t}}{\alpha^k_{r_{k+1}} +\mu}   \frac{  w^k_{n_0,r_{k+1}} h^k_{r_{k+1},t}}{\Lambda^k_{n_0,t} }  \frac{   m_{n_0,t}}{\Lambda^k_{n_0,t} }   ,\quad \text{for}\; r_{k+1} \neq r_k
\end{align}

The split in positive and negative parts is as follows for $\frac{\partial h^{k+1}_{r_{k+1},t}}{\partial h^k_{r_k,t}}$:

For $r_{k+1} = r_k$: \begin{align}
\left[\frac{\partial h^{k+1}_{r_{k+1},t}}{\partial h^k_{r_k,t}}\right]_+ &= \frac{ 1 } {\displaystyle \alpha^k_{r_k} + \mu}
\sum_n w^k_{n,r_k} \frac{m_{n,t}}{\Lambda^k_{n,t}}  ,  \\
\left[\frac{\partial h^{k+1}_{r_{k+1},t}}{\partial h^k_{r_k,t}}\right]_- &= \frac{ h^k_{r_k,t} }{\displaystyle \alpha^k_{r_k} +\mu}  \sum_n   (w^k_{n,r_k})^2 \frac{ m_{n,t}}{ (\Lambda^k_{n,t})^2 }.
\end{align}

For $r_{k+1} \neq r_k$: \begin{align}
\left[\frac{\partial h^{k+1}_{r_{k+1},t}}{\partial h^k_{r_k,t}}\right]_+ &=  0, \\
\left[\frac{\partial h^{k+1}_{r_{k+1},t}}{\partial h^k_{r_k,t}}\right]_- &=  \frac{ h^k_{r_{k+1},t} } {\displaystyle \alpha^k_{r_{k+1}}+\mu}   \displaystyle \sum_n  w^k_{n,r_{k+1}}  w^k_{n,r_k}  \frac{  m_{n,t}}{ (\Lambda_{n,t}^k)^2 } .
\end{align}

The split in positive and negative parts is as follows for $\frac{\partial h^{k+1}_{r_{k+1},t}}{\partial w^k_{n_0, r_k}}$:

For $r_{k+1} = r_k$:
\begin{align}
\left[\frac{\partial h^{k+1}_{r_{k+1},t}}{\partial w^k_{n_0, r_k}}\right]_+ &= \frac{h^k_{r_k,t}}{\alpha^k_{r_k} +\mu} \left( 1 - \frac{w^k_{n_0,r_k}    h^k_{r_k,t}}{\Lambda^k_{n_0,t}} \right) \frac{m_{n_0,t}}{ \Lambda^k_{n_0,t} },  \\
\left[\frac{\partial h^{k+1}_{r_{k+1},t}}{\partial w^k_{n_0, r_k}}\right]_- &= \frac{ h^k_{r_k,t}}{\displaystyle( \alpha^k_{r_k} +\mu )^2} \left( \sum_n  w^k_{n,r_k} \frac{m_{n,t}}{ \Lambda^k_{n,t} }  \right).
\end{align}

For $r_{k+1} \neq r_k$: \begin{align}
\left[\frac{\partial h^{k+1}_{r_{k+1},t}}{\partial w^k_{n_0, r_k}}\right]_+ &=  0, \\
\left[\frac{\partial h^{k+1}_{r_{k+1},t}}{\partial w^k_{n_0, r_k}}\right]_- &=  h^k_{r_k,t}  \frac{   m_{n_0,t}}{(\Lambda_{n_0,t}^k)^2 }  \frac{  w^k_{n_0,r_{k+1}} h^k_{r_{k+1},t}}{ \alpha^k_{r_{k+1}} +\mu }  .
\end{align}

As we will see in Section \ref{sec:mult_nmf_nonorm}, thanks to back-propagation, we never have to store these quantities in full.

\subsection{Multiplicative back-propagation updates}
\label{sec:mult-backprop}
In NMF, multiplicative updates are often derived using a heuristic approach which splits the gradient of the objective function with respect to the variable of interest into positive and negative parts, and uses the ratio of the negative part to the positive part as a multiplication factor to update the value of that variable of interest, e.g., 
\begin{align}
\bW^{k+1} = \bW^{k} \circ \frac{\left[ \nabla_{\bW} \mathcal{E} \right]_-} {\left[ {\nabla_{\bW} \mathcal{E}} \right]_+}.
\end{align}
The ratio and multiplication operations are performed elementwise.

In order to obtain multiplicative updates in our setting, we need to recursively compute the positive and negative parts of the gradient with respect to each variable.

\subsubsection{General case}

The whole gradient can be split into positive and negative terms recursively. For example, for
\begin{equation}
\frac{\partial \mathcal{E}}{\partial h^{k}_{r_k,t}} = \sum_{r_{k+1}} \frac{\partial \mathcal{E}}{\partial h^{k+1}_{r_{k+1},t}} \frac{\partial h^{k+1}_{r_{k+1},t} }{\partial h^{k}_{r_k,t}},
\end{equation}
where $h^{k}_{r_k,t}$ are the activation coefficients at time $t$ for the $r_k$th basis set in the $k$th layer, then we can compute positive and negative terms using:
\begin{align}
\left[
	\frac{\partial \mathcal{E}}
	{\partial h^{k}_{r_k,t}}
\right]_+ &= 
\sum_{r_{k+1}} 
\left( 
	\left[
		\frac{\partial \mathcal{E}}
		{\partial h^{k+1}_{r_{k+1},t}} 
	\right]_+
	\left[
		\frac{\partial h^{k+1}_{r_{k+1},t} }
		{\partial h^{k}_{r_{k},t}} 
	\right]_+
	+	
	\left[
		\frac{\partial \mathcal{E}}
		{\partial h^{k+1}_{r_{k+1},t}} 
	\right]_-
	\left[
		\frac{\partial h^{k+1}_{r_{k+1},t} }
		{\partial h^{k}_{r_k,t}}
	\right]_- 
\right), \\
\left[
	\frac{\partial \mathcal{E}}
	{\partial h^{k}_{r_k,t}}
\right]_- &= 
\sum_{r_{k+1}} 
\left( 
	\left[
		\frac{\partial \mathcal{E}}
		{\partial h^{k+1}_{r_{k+1},t}} 
	\right]_+
	\left[
		\frac{\partial h^{k+1}_{r_{k+1},t} }
		{\partial h^{k}_{r_{k},t}} 
	\right]_-
	+	
	\left[
		\frac{\partial \mathcal{E}}
		{\partial h^{k+1}_{r_{k+1},t}} 
	\right]_-
	\left[
		\frac{\partial h^{k+1}_{r_{k+1},t} }
		{\partial h^{k}_{r_k,t}}
	\right]_+ 
\right).
\end{align}
For $\frac{\partial \mathcal{E}}{\partial w^{k}_{f_k}}$, the expression involves $\frac{\partial \mathcal{E}}{\partial h^{k+1}_{r_{k+1},t}}$ and $\frac{\partial h^{k+1}_{r_{k+1},t} }{\partial w^{k}_{f_k}}$,
\begin{align}
\frac{\partial \mathcal{E}}{\partial w^{k}_{q_k}} & = \sum_t \sum_{r_{k+1}} \frac{\partial \mathcal{E}}{\partial h^{k+1}_{r_{k+1},t}} \frac{\partial h^{k+1}_{r_{k+1},t} }{\partial w^{k}_{f_k}}.
\end{align}
where $w^{k}_{f,r}$ are the values of the $r$th basis vector in the $f$th feature dimension in the $k$th layer, then the positve and negative parts are:
\begin{align}
\left[
	\frac{\partial \mathcal{E}}
	{\partial w^{k}_{f,r}} 
\right]_+ &=
\sum_t 
\sum_{r_{k+1}} 
\Bigg( 
	\left[ 
		\frac{\partial \mathcal{E}}
		{\partial h^{k+1}_{r_{k+1},t}}
	\right]_+ 
	\left[
		\frac{\partial h^{k+1}_{r_{k+1},t} }
		{\partial w^{k}_{f,r}}
	\right]_+ 
	+ 
	\left[ 
		\frac{\partial \mathcal{E}}
		{\partial h^{k+1}_{r_{k+1},t}}
	\right]_- 
	\left[
		\frac{\partial h^{k+1}_{r_{k+1},t} }
		{\partial w^{k}_{f,r}}
	\right]_- 
\Bigg)\\
\left[
	\frac{\partial \mathcal{E}}
	{\partial w^{k}_{f,r}} 
\right]_- &=
\sum_t 
\sum_{r_{k+1}} 
\Bigg( 
	\left[ 
		\frac{\partial \mathcal{E}}
		{\partial h^{k+1}_{r_{k+1},t}}
	\right]_+ 
	\left[
		\frac{\partial h^{k+1}_{r_{k+1},t} }
		{\partial w^{k}_{f,r}}
	\right]_- 
	+ 
	\left[ 
		\frac{\partial \mathcal{E}}
		{\partial h^{k+1}_{r_{k+1},t}}
	\right]_- 
	\left[
		\frac{\partial h^{k+1}_{r_{k+1},t} }
		{\partial w^{k}_{f,r}}
	\right]_+ 
\Bigg).
\end{align}
Note that this general formulation can be applied to any model with non-negative parameters, even though here we use the NMF variable names.
\subsubsection{Details for deep NMF}
\label{sec:mult_nmf_nonorm}

{\bf Splitting the gradient for $\bH$:}
\begin{align}
\frac{\partial \mathcal{E}}{\partial w^{k}_{n,r}} & = \sum_t \sum_{r_{k+1}} \frac{\partial \mathcal{E}}{\partial h^{k+1}_{r_{k+1},t}} \frac{\partial h^{k+1}_{r_{k+1},t} }{\partial w^{k}_{n,r}},
\end{align}
then:
{\allowdisplaybreaks
\begin{align}
\left[\frac{\partial \mathcal{E}}{\partial w^{k}_{n,r}} \right]_+ =&
\sum_t \sum_{r_{k+1}} \Bigg( \left[\frac{\partial h^{k+1}_{r_{k+1},t} }{\partial w^{k}_{n,r}}\right]_+ \left[ \frac{\partial \mathcal{E}}{\partial h^{k+1}_{r_{k+1},t}}\right]_+ + \left[\frac{\partial h^{k+1}_{r_{k+1},t} }{\partial w^{k}_{n,r}}\right]_- \left[ \frac{\partial \mathcal{E}}{\partial h^{k+1}_{r_{k+1},t}}\right]_- \Bigg) \notag\\
=&  \sum_t \left[\frac{\partial h^{k+1}_{r,t} }{\partial w^{k}_{n,r}}\right]_+ \left[ \frac{\partial \mathcal{E}}{\partial h^{k+1}_{r,t}}\right]_+ + \sum_t \sum_{r_{k+1}}  \left[\frac{\partial h^{k+1}_{r_{k+1},t} }{\partial w^{k}_{n,r}}\right]_- \left[ \frac{\partial \mathcal{E}}{\partial h^{k+1}_{r_{k+1},t}}\right]_-  \notag\\
=&
\sum_t \frac{h^k_{r,t}}{\alpha^k_{r} +\mu} \left( 1 - \frac{w^k_{n,r}    h^k_{r,t}}{\Lambda^k_{n,t}} \right) \frac{m_{n,t}}{ \Lambda^k_{n,t} } \left[ \frac{\partial \mathcal{E}}{\partial h^{k+1}_{r,t}}\right]_+ \notag \\
& + \sum_t \sum_{r_{k+1}} h^k_{r,t}  \frac{   m_{n,t}}{(\Lambda_{n,t}^k)^2 }  \frac{  w^k_{n,r_{k+1}} h^k_{r_{k+1},t}}{ \alpha^k_{r_{k+1}} +\mu }  \left[ \frac{\partial \mathcal{E}}{\partial h^{k+1}_{r_{k+1},t}}\right]_-  \notag \\
& - \sum_t h^k_{r,t}  \frac{   m_{n,t}}{(\Lambda_{n,t}^k)^2 }  \frac{  w^k_{n,r} h^k_{r,t}}{ \alpha^k_{r} +\mu }  \left[ \frac{\partial \mathcal{E}}{\partial h^{k+1}_{r,t}}\right]_-  \quad \text{(to compensate for the term } r_{k+1}=r_k \text{)} \notag \\
& + \sum_t \frac{ h^k_{r,t}}{\displaystyle( \alpha^k_{r} +\mu )^2} \left( \sum_{n'}  w^k_{n',r} \frac{m_{n',t}}{ \Lambda^k_{n',t} }  \right) \left[ \frac{\partial \mathcal{E}}{\partial h^{k+1}_{r,t}}\right]_- \quad \text{(same)}\notag \\
=&
\sum_t  \frac{m_{n,t}}{ \Lambda^k_{n,t} } \left( \frac{h^k_{r,t}}{\alpha^k_{r} +\mu}\left[ \frac{\partial \mathcal{E}}{\partial h^{k+1}_{r,t}}\right]_+ \right)
-
w^k_{n,r} \sum_t  \frac{m_{n,t}}{ (\Lambda^k_{n,t})^2 } \left( \frac{(h^k_{r,t})^2}{\alpha^k_{r} +\mu}  \left[ \frac{\partial \mathcal{E}}{\partial h^{k+1}_{r,t}}\right]_+ \right)
\notag \\
& + \sum_t h^k_{r,t} \left\{ \frac{   m_{n,t}}{(\Lambda_{n,t}^k)^2 }
\sum_{r_{k+1}} \frac{  w^k_{n,r_{k+1}} }{ \alpha^k_{r_{k+1}} +\mu }
\left( h^k_{r_{k+1},t} \left[ \frac{\partial \mathcal{E}}{\partial h^{k+1}_{r_{k+1},t}}\right]_- \right) \right\} \notag \\
& - w^k_{n,r} \sum_t  \frac{   m_{n,t}}{(\Lambda_{n,t}^k)^2 } \left( \frac{ (h^k_{r,t})^2 }{ \alpha^k_{r} +\mu }  \left[ \frac{\partial \mathcal{E}}{\partial h^{k+1}_{r,t}}\right]_- \right) \notag \\
& + \sum_t \frac{ h^k_{r,t}}{\displaystyle( \alpha^k_{r} +\mu )^2} \left( \sum_{n'}  w^k_{n',r} \frac{m_{n',t}}{ \Lambda^k_{n',t} }  \right) \left[ \frac{\partial \mathcal{E}}{\partial h^{k+1}_{r,t}}\right]_-.
\end{align}
It is very important to carefully compute these quantities to avoid having to consider computations or storage involving tensors. It turns out that the above expression can be reformulated using only matrix operations.
Altogether, we get:
\begin{align}
\left[\nabla_{\bW^{k}}\mathcal{E} \right]_+ =&
 \frac{\bM}{ \bLambda^k } \left( \frac{\bH^k}{(\bW^k)^\intercal \mathbf{1}_{N \times T} +\mu} \circ \left[ \nabla_{\bH^{k+1}}\mathcal{E} \right]_+ \right)^\intercal \notag \\
& + \left\{ \frac{   \bM }{(\bLambda^k)^2 } \circ \left(
\frac{  \bW^k }{ (\bW^k)^\intercal \mathbf{1}_{N \times R}  + \mu }
\left( \bH^k \circ \left[ \nabla_{\bH^{k+1}}\mathcal{E} \right]_- \right) \right) \right\} (\bH^k)^\intercal \notag \\
& - \bW^k \circ \left\{ \frac{ \bM }{(\bLambda^k)^2 } \left( \frac{ (\bH^k)^2 \circ \left( \left[ \nabla_{\bH^{k+1}}\mathcal{E}\right]_+ + \left[ \nabla_{\bH^{k+1}}\mathcal{E}\right]_- \right)}{ (\bW^k)^\intercal \mathbf{1}_{N \times T} +\mu }  \right)^\intercal \right\} \notag \\
& +\frac{ \mathbf{1}_{R \times T}  \left( \bH^k \circ  \left( (\bW^k)^\intercal \frac{\bM}{ \bLambda^k }  \right) \circ \left[ \nabla_{\bH^{k+1}}\mathcal{E} \right]_- \right)^\intercal }{\displaystyle( (\bW^k)^\intercal \mathbf{1}_{N \times R} +\mu )^2} ,
\end{align}
}
where $\mathbf{1}_{a \times b}$ is an $a \times b$ matrix of all $1$. This matrix is only used here for notation purposes, and in practice, we use MATLAB's bsxfun to avoid having to explicitely create it.

The negative part of the gradient is exactly the same, except that $ \left[\frac{\partial \mathcal{E}}{\partial h^{k+1}_{r_{k+1},t}}\right]_- $ and $ \left[\frac{\partial \mathcal{E}}{\partial h^{k+1}_{r_{k+1},t}}\right]_+ $ are interchanged:
\begin{align}
\left[\frac{\partial \mathcal{E}}{\partial w^{k}_{n,r}} \right]_- =&
\sum_t \sum_{r_{k+1}} \Bigg( \left[\frac{\partial h^{k+1}_{r_{k+1},t} }{\partial w^{k}_{n,r}}\right]_+ \left[ \frac{\partial \mathcal{E}}{\partial h^{k+1}_{r_{k+1},t}}\right]_- + \left[\frac{\partial h^{k+1}_{r_{k+1},t} }{\partial w^{k}_{n,r}}\right]_- \left[ \frac{\partial \mathcal{E}}{\partial h^{k+1}_{r_{k+1},t}}\right]_+ \Bigg) \notag\\
=&  \sum_t \left[\frac{\partial h^{k+1}_{r,t} }{\partial w^{k}_{n,r}}\right]_+ \left[ \frac{\partial \mathcal{E}}{\partial h^{k+1}_{r,t}}\right]_- + \sum_t \sum_{r_{k+1}}  \left[\frac{\partial h^{k+1}_{r_{k+1},t} }{\partial w^{k}_{n,r}}\right]_- \left[ \frac{\partial \mathcal{E}}{\partial h^{k+1}_{r_{k+1},t}}\right]_+,
\end{align}
\begin{align}
\left[\nabla_{\bW^{k}}\mathcal{E} \right]_- =&
 \frac{\bM}{ \bLambda^k } \left( \frac{\bH^k}{(\bW^k)^\intercal \mathbf{1}_{N \times T} +\mu} \circ \left[ \nabla_{\bH^{k+1}}\mathcal{E} \right]_- \right)^\intercal \notag \\
& + \left\{ \frac{   \bM }{(\bLambda^k)^2 } \circ \left(
\frac{  \bW^k }{ (\bW^k)^\intercal \mathbf{1}_{N \times R}  + \mu }
\left( \bH^k \circ \left[ \nabla_{\bH^{k+1}}\mathcal{E} \right]_+ \right) \right) \right\} (\bH^k)^\intercal \notag \\
& - \bW^k \circ \left\{ \frac{ \bM }{(\bLambda^k)^2 } \left( \frac{ (\bH^k)^2 \circ \left( \left[ \nabla_{\bH^{k+1}}\mathcal{E}\right]_+ + \left[ \nabla_{\bH^{k+1}}\mathcal{E}\right]_- \right)}{ (\bW^k)^\intercal \mathbf{1}_{N \times T} +\mu }  \right)^\intercal \right\} \notag \\
& +\frac{ \mathbf{1}_{R \times T}  \left( \bH^k \circ  \left( (\bW^k)^\intercal \frac{\bM}{ \bLambda^k }  \right) \circ \left[ \nabla_{\bH^{k+1}}\mathcal{E} \right]_+ \right)^\intercal }{\displaystyle( (\bW^k)^\intercal \mathbf{1}_{N \times R} +\mu )^2}.
\end{align}

{\bf Splitting the gradient for $\bH^k$:}
Now the gradient with respect to $\bH^k$, which is used recursively:
\begin{align}
\left[\frac{\partial \mathcal{E}}{\partial h^{k}_{r_k,t}}\right]_+ &=  \left[\frac{\partial \mathcal{E}}{\partial h^{k+1}_{r_{k},t}} \right]_+
\left[\frac{\partial h^{k+1}_{r_{k},t} }{\partial h^{k}_{r_k,t}} \right]_+
+ \sum_{r_{k+1}}
\left[\frac{\partial \mathcal{E}}{\partial h^{k+1}_{r_{k+1},t}} \right]_-
\left[\frac{\partial h^{k+1}_{r_{k+1},t} }{\partial h^{k}_{r_k,t}}\right]_-, \notag\\
&=  \left[\frac{\partial \mathcal{E}}{\partial h^{k+1}_{r_{k},t}} \right]_+
\left[\frac{\partial h^{k+1}_{r_{k},t} }{\partial h^{k}_{r_k,t}} \right]_+
+ \sum_{r_{k+1}}
\left[\frac{\partial \mathcal{E}}{\partial h^{k+1}_{r_{k+1},t}} \right]_-
\frac{ h^k_{r_{k+1},t} } {\displaystyle \alpha^k_{r_{k+1}}+\mu}   \displaystyle \sum_n  w^k_{n,r_{k+1}}  w^k_{n,r_k}  \frac{  m_{n,t}}{ (\Lambda_{n,t}^k)^2 } \notag \\
&=  \left[\frac{\partial \mathcal{E}}{\partial h^{k+1}_{r_{k},t}} \right]_+
\left[\frac{\partial h^{k+1}_{r_{k},t} }{\partial h^{k}_{r_k,t}} \right]_+
+ \sum_n w^k_{n,r_k}  \left\{ \frac{  m_{n,t}}{ (\Lambda_{n,t}^k)^2 } \left( \sum_{r_{k+1}} \Big(
\left[\frac{\partial \mathcal{E}}{\partial h^{k+1}_{r_{k+1},t}} \right]_-
 h^k_{r_{k+1},t}   \Big) \frac{w^k_{n,r_{k+1}} } {\displaystyle \alpha^k_{r_{k+1}}+\mu} \right) \right\},
\end{align}
where
\begin{align}
\left[\frac{\partial h^{k+1}_{r_k,t}}{\partial h^k_{r_k,t}}\right]_+ &= \frac{ 1 } {\displaystyle \alpha^k_{r_k} + \mu}
\sum_n w^k_{n,r_k} \frac{m_{n,t}}{\Lambda^k_{n,t}}.
\end{align}

This can also be rewritten using matrix computations:
\begin{align}
\left[\nabla_{\bH^{k}}\mathcal{E}\right]_+ &= \frac{ \left[\nabla_{\bH^{k+1}}\mathcal{E} \right]_+ \circ
\left( (\bW^k)^\intercal \frac{\bM}{ \bLambda^k }  \right) } { (\bW^k)^\intercal \mathbf{1}_{N \times T} +\mu }
+
(\bW^k )^\intercal
\left\{ \frac{ \bM}{ (\bLambda^k)^2 } \circ
\left(
\bW^k
    \big(
    \frac{ \left[\nabla_{\bH^{k+1}}\mathcal{E} \right]_- \circ \bH^k}
         {(\bW^k)^\intercal \mathbf{1}_{N \times T} +\mu }
    \big)
\right)
\right\}.
\end{align}

The negative part is similar again, just interchanging $\left[\nabla_{\bH^{k+1}}\mathcal{E} \right]_+$ and $\left[\nabla_{\bH^{k+1}}\mathcal{E} \right]_-$:
\begin{align}
\left[\frac{\partial \mathcal{E}}{\partial h^{k}_{r_k,t}}\right]_- &=
\left[\frac{\partial \mathcal{E}}{\partial h^{k+1}_{r_{k},t}} \right]_-
\left[\frac{\partial h^{k+1}_{r_{k},t} }{\partial h^{k}_{r_k,t}}\right]_+
+
\sum_{r_{k+1}} \left( \left[\frac{\partial \mathcal{E}}{\partial h^{k+1}_{r_{k+1},t}} \right]_+
\left[\frac{\partial h^{k+1}_{r_{k+1},t} }{\partial h^{k}_{r_k,t}} \right]_- \right),
\end{align}
and in matrix notations:
\begin{align}
\left[\nabla_{\bH^{k}}\mathcal{E}\right]_- &=
\frac{ \left[\nabla_{\bH^{k+1}}\mathcal{E} \right]_- \circ \left( (\bW^k)^\intercal \frac{\bM}{ \bLambda^k }  \right) }
     { (\bW^k)^\intercal \mathbf{1}_{N \times T} +\mu }
+
(\bW^k )^\intercal
\left\{ \frac{ \bM}{ (\bLambda^k)^2 } \circ
\left(
\bW^k
    \big(
    \frac{ \left[\nabla_{\bH^{k+1}}\mathcal{E} \right]_+ \circ \bH^k}
         {(\bW^k)^\intercal \mathbf{1}_{N \times T} +\mu }
    \big)
\right)
\right\}.
\end{align}

\subsection{Deep NMF gradient computation procedure}
Let us put everything together for the case of single-source reconstruction with Wiener filter and minimization of L2 distance ($\beta=2$) at the last layer, and KL-divergence NMF updates for $\bH^k$ without normalization of $\bW^k$ in the layers where we optimize $\bW^k$ (but with normalization for all layers up to there). Assuming that a forward computation of $\bH^k$ for $k=1,\ldots,K$ as been performed, we compute positive and negative parts of the gradients as follows:
\begin{itemize}
\item Compute the gradient for the last layer with respect to $\bH^K$.
\begin{align}
\left[\nabla_{\bH^{K,l}}{\mathcal{E}}\right]_+ &=  {{\bW}^{K,l}}^\intercal
         \frac{ {\bf M}^2  \circ {\bf \Lambda}^l \circ {\bf \Lambda}^{\overline{l}} } { {\bf \Lambda}^3 },  \\
\left[\nabla_{\bH^{K,l}}{\mathcal{E}}\right]_- &= {{\bW}^{K,l}}^\intercal
         \frac{ {\bf M} \circ {\bf S}^l  \circ {\bf \Lambda}^{\overline{l}} } { {\bf \Lambda}^2 },   \\
\left[\nabla_{\bH^{K,\overline{l}}}{\mathcal{E}}\right]_+ &= {{\bW}^{K,\overline{l}}}^\intercal
         \frac{ {\bf M} \circ {\bf S}^l \circ {\bf \Lambda}^l } { {\bf \Lambda}^2 } , \\
\left[\nabla_{\bH^{K,\overline{l}}}{\mathcal{E}}\right]_- &= {{\bW}^{K,\overline{l}}}^\intercal
         \frac{ {\bM}^2  \circ ({\bf \Lambda}^l)^2 } { {\bf \Lambda}^3 },
\end{align}
where $\bLambda = \bLambda^K$, $\bLambda^l = \bLambda^{K,l}$, $\bLambda^{\overline{l}} = \bLambda^{K,\overline{l}}$.
\item Compute recursively the gradient for lower layers with respect to $\bH^k$ using:
\begin{align}
\left[\nabla_{\bH^{k}}\mathcal{E}\right]_+ &= \frac{
\left( (\bW^k)^\intercal \frac{\bM}{ \bLambda^k } \right) \circ  \left[\nabla_{\bH^{k+1}}\mathcal{E} \right]_+  } { (\bW^k)^\intercal \mathbf{1}_{N \times T} +\mu } \notag \\
& \hspace{2cm} 
+
(\bW^k )^\intercal
\left\{ \frac{ \bM}{ (\bLambda^k)^2 } \circ
\left(
    \frac{\bW^k }
         { (\bW^k)^\intercal \mathbf{1}_{N \times R}  + \mu }
    \left( \bH^k \circ \left[ \nabla_{\bH^{k+1}}\mathcal{E} \right]_- \right)
\right)
\right\}  \\
\left[\nabla_{\bH^{k}}\mathcal{E}\right]_- &=
\frac{
\left( (\bW^k)^\intercal \frac{\bM}{ \bLambda^k } \right) \circ  \left[\nabla_{\bH^{k+1}}\mathcal{E} \right]_-  } { (\bW^k)^\intercal \mathbf{1}_{N \times T} +\mu } \notag \\
& \hspace{2cm} 
+
(\bW^k )^\intercal
\left\{ \frac{ \bM}{ (\bLambda^k)^2 } \circ
\left(
    \frac{\bW^k }
         { (\bW^k)^\intercal \mathbf{1}_{N \times R}  + \mu }
    \left( \bH^k \circ \left[ \nabla_{\bH^{k+1}}\mathcal{E} \right]_+ \right)
\right)
\right\}.
\end{align}
\item Compute the gradient with respect to $\bW^k$ for those layers where we want to optimize $\bW^k$. If $\bW^k$ is tied across layers, the gradients at those layers can simply be summed.
{\allowdisplaybreaks
\begin{align}
\left[\nabla_{\bW^{k}}\mathcal{E} \right]_+ =&
 \frac{\bM}{ \bLambda^k } \left( \frac{\bH^k \circ \left[ \nabla_{\bH^{k+1}}\mathcal{E} \right]_+}{(\bW^k)^\intercal \mathbf{1}_{N \times T} +\mu}  \right)^\intercal \notag \\
& + \left\{ \frac{   \bM }{(\bLambda^k)^2 } \circ \left(
\frac{  \bW^k }{ (\bW^k)^\intercal \mathbf{1}_{N \times R}  + \mu }
\left( \bH^k \circ \left[ \nabla_{\bH^{k+1}}\mathcal{E} \right]_- \right) \right) \right\} (\bH^k)^\intercal \notag \\
& - \bW^k \circ \left\{ \frac{ \bM }{(\bLambda^k)^2 } \left( \frac{ (\bH^k)^2 \circ \left( \left[ \nabla_{\bH^{k+1}}\mathcal{E}\right]_+ + \left[ \nabla_{\bH^{k+1}}\mathcal{E}\right]_- \right)}{ (\bW^k)^\intercal \mathbf{1}_{N \times T} +\mu }  \right)^\intercal \right\} \notag \\
& +\frac{ \mathbf{1}_{R \times T}  \left( \bH^k \circ  \left( (\bW^k)^\intercal \frac{\bM}{ \bLambda^k }  \right) \circ \left[ \nabla_{\bH^{k+1}}\mathcal{E} \right]_- \right)^\intercal }{\displaystyle( (\bW^k)^\intercal \mathbf{1}_{N \times R} +\mu )^2} \\
\left[\nabla_{\bW^{k}}\mathcal{E} \right]_- =&
 \frac{\bM}{ \bLambda^k } \left( \frac{\bH^k \circ \left[ \nabla_{\bH^{k+1}}\mathcal{E} \right]_-}{(\bW^k)^\intercal \mathbf{1}_{N \times T} +\mu}  \right)^\intercal \notag \\
& + \left\{ \frac{   \bM }{(\bLambda^k)^2 } \circ \left(
\frac{  \bW^k }{ (\bW^k)^\intercal \mathbf{1}_{N \times R}  + \mu }
\left( \bH^k \circ \left[ \nabla_{\bH^{k+1}}\mathcal{E} \right]_+ \right) \right) \right\} (\bH^k)^\intercal \notag \\
& - \bW^k \circ \left\{ \frac{ \bM }{(\bLambda^k)^2 } \left( \frac{ (\bH^k)^2 \circ \left( \left[ \nabla_{\bH^{k+1}}\mathcal{E}\right]_+ + \left[ \nabla_{\bH^{k+1}}\mathcal{E}\right]_- \right)}{ (\bW^k)^\intercal \mathbf{1}_{N \times T} +\mu }  \right)^\intercal \right\} \notag \\
& +\frac{ \mathbf{1}_{R \times T}  \left( \bH^k \circ  \left( (\bW^k)^\intercal \frac{\bM}{ \bLambda^k }  \right) \circ \left[ \nabla_{\bH^{k+1}}\mathcal{E} \right]_+ \right)^\intercal }{\displaystyle( (\bW^k)^\intercal \mathbf{1}_{N \times R} +\mu )^2}.
\end{align}
}
\item From there, we can form multiplicative update equations for $\bW^k$.
\end{itemize}

Note that only the $\left[ \nabla_{\bH^{k+1}}\mathcal{E} \right]_\pm$ for the layer $k+1$ need to be available when performing computations for $\bH^k$ and $\bW^k$, and they can thus be replaced iteratively as we descend through layers, limiting the memory requirements.

\end{document}